\documentclass[letterpaper, 10 pt, conference]{ieeeconf}  

\IEEEoverridecommandlockouts                              

\usepackage{graphics} 
\usepackage{epsfig} 
\usepackage{color}
\usepackage{times} 
\usepackage{amsmath} 
\usepackage{amssymb}  

\usepackage{algorithm}
\usepackage[noEnd=false]{algpseudocodex}
\usepackage{tikz}
\usetikzlibrary{arrows.meta, positioning, shapes.geometric, fit, calc, backgrounds}
\usepackage{tabularx}
\usepackage[hidelinks]{hyperref}
\usepackage{booktabs}
\usepackage{subcaption}
\usepackage[font=small]{caption}
\usepackage{siunitx}
\newcommand{\projectname}{Hierarchical RL-Diffusion Policy for Efficient Nonprehensile Manipulation}
\newcommand{\hook}{Push Smarter, Not Harder: }

\title{\LARGE \bf
\hook \projectname
}

\author{Steven Caro, Stephen L. Smith
\thanks{Resources used in preparing this research were provided, in part, by the Province of Ontario, the Government of Canada through CIFAR, and companies sponsoring the Vector Institute.}
\thanks{The authors are with the Department of Electrical and Computer Engineering, University of Waterloo, Waterloo, ON N2L 3G1, Canada (e-mails:\protect\url{{steven.caro, stephen.smith}@uwaterloo.ca})}
}

\begin{document}

\maketitle
\thispagestyle{empty}
\pagestyle{empty}

\setlength{\belowcaptionskip}{-5pt}

\begin{abstract}
\label{sec:abstract}
Nonprehensile manipulation, such as pushing objects across cluttered environments, presents a challenging control problem due to complex contact dynamics and long-horizon planning requirements. In this work, we propose HeRD, a hierarchical reinforcement learning-diffusion policy that decomposes
pushing tasks
into two levels: high-level goal selection and low-level trajectory generation. We employ a high-level reinforcement learning (RL) agent to select intermediate spatial goals, and a low-level goal-conditioned diffusion model to generate feasible, efficient trajectories to reach them.

This architecture combines the 
long-term reward maximizing behaviour of RL with the generative capabilities of diffusion models.
We evaluate our method in a  2D simulation environment and show that it outperforms
the state-of-the-art baseline
in success rate, path efficiency, and generalization across multiple environment configurations. Our results suggest that hierarchical control with generative low-level planning is a promising direction for scalable, goal-directed nonprehensile manipulation. Code, documentation, and trained models are available:
\url{https://github.com/carosteven/HeRD}.
\end{abstract}
\section{Introduction}
\label{sec:introduction}

Nonprehensile manipulation (manipulation without grasping) is a fundamental capability for general-purpose robots operating in cluttered or constrained environments. Compared to prehensile manipulators, such as gripper arms, nonprehensile manipulators can interact with multiple objects and are more accommodating of simple, low-cost hardware. While gripper arms offer some unique advantages, many real-world scenarios, such as area-clearing or transporting large or ungraspable objects, are better suited for nonprehensile strategies. Although humans intuitively learn these skills at a young age, the task remains highly complex in robotics due to the nonlinear and contact-rich dynamics of pushing \cite{stuber_lets_2020}.

Reinforcement learning (RL) is well-suited for goal-directed tasks, as it optimizes behaviour through trial and error based on reward signals. In robotic settings, RL policies often operate over low-level control signals such as wheel velocities or discrete movement commands (e.g., move forward, turn left) \cite{aljalbout_role_2024}. As a result, these policies must simultaneously learn task-relevant strategies and develop an understanding of the robot’s dynamics. This dual burden makes learning more challenging, especially in contact-rich tasks like nonprehensile manipulation. Hierarchical RL addresses this challenge by separating high-level decision-making from low-level control. Wu et al. demonstrated that using Spatial Action Maps (SAM) \cite{wu_spatial_2020}, where an RL policy outputs spatial goals, combined with a separate controller for execution, improves performance in nonprehensile manipulation tasks.

Diffusion models are a branch of generative models originally developed for image generation tasks, but have recently shown promising applications in robotics due to their ability to produce diverse, high-dimensional, and temporally coherent trajectories \cite{janner_planning_2022, chi_diffusion_2024}. Trained on human demonstrations, diffusion policies capture rich behavioural priors that reflect intuitive control strategies. In our case, these include
stabilizing contact during object manipulation, anticipating future interactions, and avoiding disruptive collisions
-- strategies which are intuitive for a human but
are difficult to learn from reinforcement alone, especially in the presence of sparse or delayed rewards. While much promise is shown here, diffusion policies are limited by the quality and diversity of the demonstrations provided.
To address this limitation, we offload high-level planning to an RL agent, allowing the diffusion policy to focus on trajectory generation, where high quality demonstrations are easier to collect at low cost.

\begin{figure}
    \centering
    \begin{minipage}{\columnwidth}
        \centering
        \foreach \i in {1,...,5}{
            \includegraphics[width=0.36\columnwidth, angle=90]{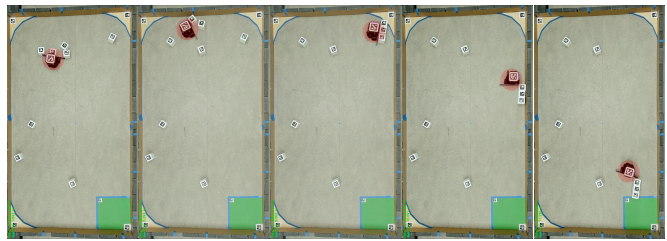} \hspace{-10pt}
        }
    \end{minipage}
    \caption{
    The robot (highlighted in red) is tasked with pushing the boxes into the green receptacle. This sequence of images shows a robot using the proposed HeRD policy gathering three boxes and pushing them into the receptacle.
    }
    \label{fig:physical}
\end{figure}

We seek to improve the spatial action maps formulation in two ways: increasing success rate and lowering the distance required of the agent to complete the task. 
Our work makes the following contributions:
\begin{itemize}
    \item We introduce a goal-conditioned diffusion policy framework capable of producing collision-free trajectories by enforcing feasibility constraints. We then train this policy using human demonstrations to generate trajectories to navigate to spatial goals.
    \item We propose \textbf{H}i\textbf{e}rarchical \textbf{R}einforcement Learning - \textbf{D}iffusion Policy (\textbf{HeRD}) that combines the high-level planning of spatially-aware RL with the context-aware trajectory generation of a diffusion policy, and outperforms the state-of-the-art SAM policy in \cite{wu_spatial_2020} in terms of success and distance in a range of environments. Key to our approach is a new reward function that encourages the RL agent to act more efficiently.
    \item Finally, in our physical implementation, we observe a substantial increase in success rate of the HeRD policy when compared to the current state-of-the-art policy.
\end{itemize}

\section{Related Work}
\label{sec:related_work}

Nonprehensile manipulation is inherently challenging due to the underactuated nature of the system from
the limited set of forces that can be transmitted to the object from an agent \cite{hogan_reactive_2020},
and the highly non-linear dynamics of pushing \cite{stuber_lets_2020}.
Prior work has addressed these challenges using model-based and feedback control approaches. For instance, reactive controllers adapt to contact uncertainties during pushing \cite{ozdamar_pushing_2024}, while model predictive control 
methods reason over contact forces and future trajectories to maintain stable pushes \cite{tang_unwieldy_2023, bauza_data-efficient_2018}.

More recently, learning-based methods, particularly reinforcement learning, have become popular for developing closed-loop manipulation policies. Action space design plays a crucial role in these methods. Continuous-control policies typically operate directly in low-level action spaces such as linear and angular velocities \cite{sun_integrating_2023, del_aguila_ferrandis_nonprehensile_2023}. While expressive, these actions combine low-level motion control with high-level planning, forcing the RL agent to learn both precise robot dynamics and long-horizon task structure simultaneously, making training inefficient. On the other hand, discretized motion primitive action spaces improve learning stability \cite{yuan_rearrangement_2018}, but 
they introduce their own challenges: meaningful reward signals may arise only after executing long sequences of motion primitives, making exploration difficult in long-horizon tasks \cite{wu_spatial_2020}.

Spatial Action Maps (SAMs) address this tradeoff by treating the action space as a dense grid of spatial goals. They decouple high-level decision-making from low-level motion control, improving sample efficiency and generalization. 
The discrete nature of SAMs lends them to be a useful action representation for Deep Q-Networks \cite{mnih_playing_2013}. In our work, we use the Double DQN variant \cite{hasselt_deep_2015}.
SAMs have shown success in pushing \cite{wu_spatial_2020}, blowing \cite{wu_learning_2022}, and multi-agent manipulation \cite{wu_spatial_2021}. However, these works do not consider the state of the environment when generating paths from the output of the SAM; we address this by using a generative diffusion policy conditioned on the state of the environment to make context-aware paths.

The SAM framework can be viewed as a form of implicit hierarchy: the high-level agent chooses a spatial subgoal, and a low-level controller executes it.
Hierarchical Reinforcement Learning (HRL) decomposes complex tasks into a hierarchy of subgoals or options
\cite{sutton_between_1999, vezhnevets_feudal_2017}.
This reduces the effective planning horizon and improves sample efficiency, particularly in sparse-reward settings.
Our method makes this hierarchy explicit by pairing a goal-predicting DDQN policy with a denoising diffusion probabilistic model (DDPM) \cite{sohl-dickstein_deep_2015, ho_denoising_2020} for path generation. DDPMs are generative diffusion models with recent applications in robotics that sample structured action sequences via iterative denoising \cite{janner_planning_2022} and can be conditioned on observations \cite{chi_diffusion_2024}.


A central challenge in diffusion-based control is guiding the generation process toward task-relevant outcomes. 
In our framework, we use goal inpainting \cite{janner_planning_2022}, which conditions the trajectory endpoints by fixing them to specific targets, as well as FiLM \cite{perez_film_2018} layers to encode observations to ensure geometry-aware trajectory synthesis.

It is worth noting related works that use hierarchical diffusion policies.
HDMI \cite{li_hierarchical_2023}, Hierarchical Diffuser \cite{chen_simple_2024}, and HDP \cite{ma_hierarchical_2024} all adopt a two-level diffusion architecture: a high-level model generates subgoals or intermediate keyframes, and a low-level diffusion model produces the full trajectory segments that reach them.
Other approaches, such as \cite{wang_hierarchical_2025, wu_diffusion-reinforcement_2025}, combine high-level diffusion models for subgoal generation with low-level controllers trained via reinforcement learning, forming hybrid architectures that integrate generative planning with policy-based execution.

Our method adopts a complementary hierarchical structure: the high-level planner is a value-based reinforcement learning policy that selects spatial subgoals, while the low-level controller is a goal-conditioned diffusion model that synthesizes smooth, feasible trajectories to those targets.

\section{Problem Formulation and Spatial Action Map Overview}
\label{sec:problemformulation}

In this section, we first formulate the task as a Markov decision process and specify the objectives we aim to optimize.
We then provide an overview of spatial action maps, including details about the state and action spaces, and the reward function used in the original paper.

\subsection{Problem Formulation}
We consider an environment $\mathcal{W} \subset \mathbb{R}^2$ containing
$n$ fixed obstacles $\mathcal{O}_1, \ldots, \mathcal{O}_n \subset \mathcal{W}$, a receptacle $\mathcal{B} \subset \mathcal{W}$, $m$ movable boxes, and a mobile robot.
At time $t \geq 0$, each box $j\in\{1,\ldots,m\}$ is described by a time-varying set $b_j(t) \subset \mathcal{W}$, defining the 2D space occupied by the box. Likewise, the robot is described by $\mathcal{R}(t) \subset \mathcal{W}$.
The robot navigates within the free space $\mathcal{W}_{\text{free}} = \mathcal{W} \setminus \left( \cup_i \mathcal{O}_i \right)$.
The task of the robot is to successfully push all boxes into the receptacle by a given time $T_{\max}$: That is, to reach a state at some time $t'\in[0,T_{\max}]$ where $b_j(t') \subset \mathcal{B}$ for all $j \in \{1, \ldots, m\}$.

\vspace{0.5em}
\noindent \textbf{Objective.}
Our objective is to compute a robot policy $\pi$ that optimizes a two-tiered objective, capturing both task success and the efficiency of robot behaviour. The \emph{Primary Objective} is to maximize the expected number of boxes placed in the receptacle by time $T_{\max}$.  Formally, the primary objective corresponds to maximizing the success indicator:
    \begin{equation*}
        \sum_{j=1}^m \mathbf{1}[b_j(T_{\max})\subset \mathcal{B}].
    \end{equation*}
The \emph{Secondary Objective} is to minimize the time to complete the task. Formally, we seek to minimize:
    \begin{equation*}
        \min \big\{t \in[0,T_{\max}] \; | \; b_j(t) \subset \mathcal{B} \text{ $\forall \; j\in\{1,\ldots,m\}$}\big\},
    \end{equation*}
    where the completion time is $T_{\max}$ if not all boxes are in the receptacle by time $T_{\max}$.

This tiered formulation encourages the agent to reliably complete the task while favouring efficient policies.

\vspace{0.5em}
\noindent\textbf{MDP Model.} We model the problem as a Markov decision process (MDP), defined by the tuple $(\mathcal{S}, \mathcal{A}, \mathbb{P}, R, \gamma)$, where
$\mathcal{S}$ is the state space consisting of the robot position and environment configuration and $\mathcal{A}$ is the action space.
At each time step, the robot observes a state $s \in \mathcal{S}$ and chooses an action $a \in \mathcal{A}$ according to a policy $\pi:\mathcal{S}\to \mathcal{A}$. The environment transitions to a new state $s' \sim \mathbb{P}(s'|s, a)$ and emits a scalar reward $r = R(s, a)$, where the scalar reward acts as a proxy for the multi-objective optimization stated above.  The constant $\gamma \in (0,1)$ is the discount factor.

\subsection{Spatial Action Maps}
\label{sec:sam_details}

In this section, we review the spatial action map (SAM) policy as proposed in~\cite{wu_spatial_2020}.
To motivate the use of SAMs, we first formulate the task as a reinforcement learning problem. 
To solve the above MDP, we adopt a value-based RL approach using a Double Deep Q-Network (DDQN) \cite{hasselt_deep_2015}. The goal is to learn an action-value function $Q_\theta:\mathcal{S} \times \mathcal{A} \rightarrow \mathbb{R}$, parameterized by $\theta$, that estimates the expected cumulative discounted reward when taking action $a$ in state $s$ and following an optimal policy thereafter.

During training, transitions $(s,a,r,s')$ are uniformly sampled from an experience replay buffer \cite{lin_self-improving_1992}, and the online network parameters $\theta$ are updated to minimize the following temporal difference loss:
\begin{equation*}
    \mathcal{L}(\theta) = \left| r + \gamma Q_{\theta'}\left( s', \arg\max_{a'} Q_\theta(s', a') \right) - Q_\theta(s, a) \right|,
\end{equation*}
where $Q_{\theta'}$ is the target network with frozen parameters $\theta'$ periodically updated from $\theta$.

The resulting policy is derived greedily from the learned Q-function:
\begin{equation*}
    \pi_{\theta}(s) = \arg\max_a Q_\theta(s, a).
\end{equation*}

SAMs define a dense, pixel-aligned action space where each action corresponds to a goal coordinate in the robot’s environment. The Q-network $Q_\theta(s, a)$ outputs a heatmap over all spatial locations, representing the value of navigating to each coordinate. The policy $\pi_{\theta}(s)$ selects the pixel with the highest Q-value, which becomes the high-level goal for the robot.
A path is then generated to this goal point using the Shortest Path Faster Algorithm (SPFA) \cite{fanding_spfa_1994}, and
a low-level controller is used to traverse the path. This process is summarized in Algorithm \ref{alg:orig}. The rest of this section details the state space, action space, and reward function, which are central to this formulation.

\begin{algorithm}
\caption{Spatial Action Map rollout (based on \cite{wu_spatial_2020})}
\label{alg:orig}
\begin{small}
\begin{algorithmic}[1]
\Require policy $\pi_{\theta}$, robot position $p_{\mathcal{R}}$, controller $c$
\State Observe state $s$
\State Select action $a = \pi_{\theta}$
\State $\texttt{path} \leftarrow \texttt{SPFA}(p_{\mathcal{R}}, a)$
\State Initialize controller $c_{\text{active}} \gets c(\texttt{path})$
\State Execute $c_{active}$ until goal $a$ is reached
\end{algorithmic}
\end{small}
\end{algorithm}

\begin{figure}[tbp]
  \centering
  \begin{subfigure}[t]{0.24\columnwidth}
    \centering
    \includegraphics[width=\linewidth]{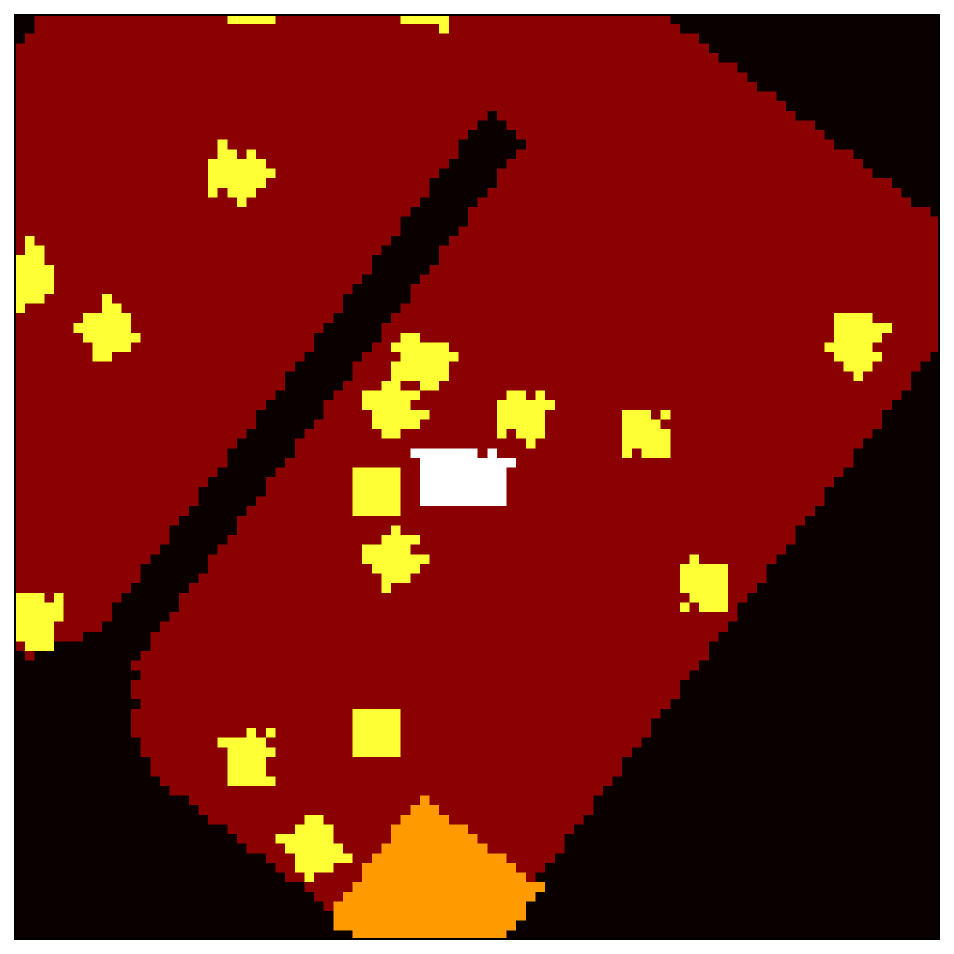}
    \caption{Overhead Map}
    \label{fig:obs_0}
  \end{subfigure}
  \begin{subfigure}[t]{0.24\columnwidth}
    \centering
    \includegraphics[width=\linewidth]{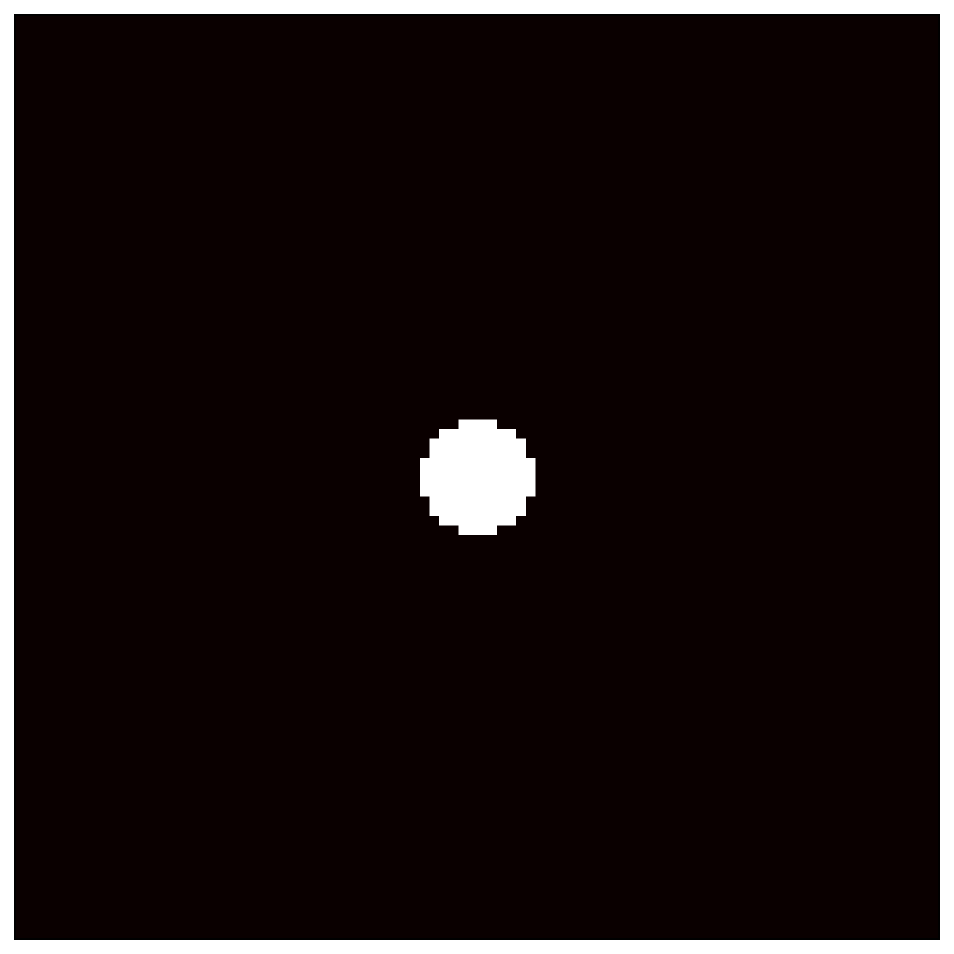}
    \caption{Robot Footprint}
    \label{fig:obs_1}
  \end{subfigure}
  \begin{subfigure}[t]{0.24\columnwidth}
    \centering
    \includegraphics[width=\linewidth]{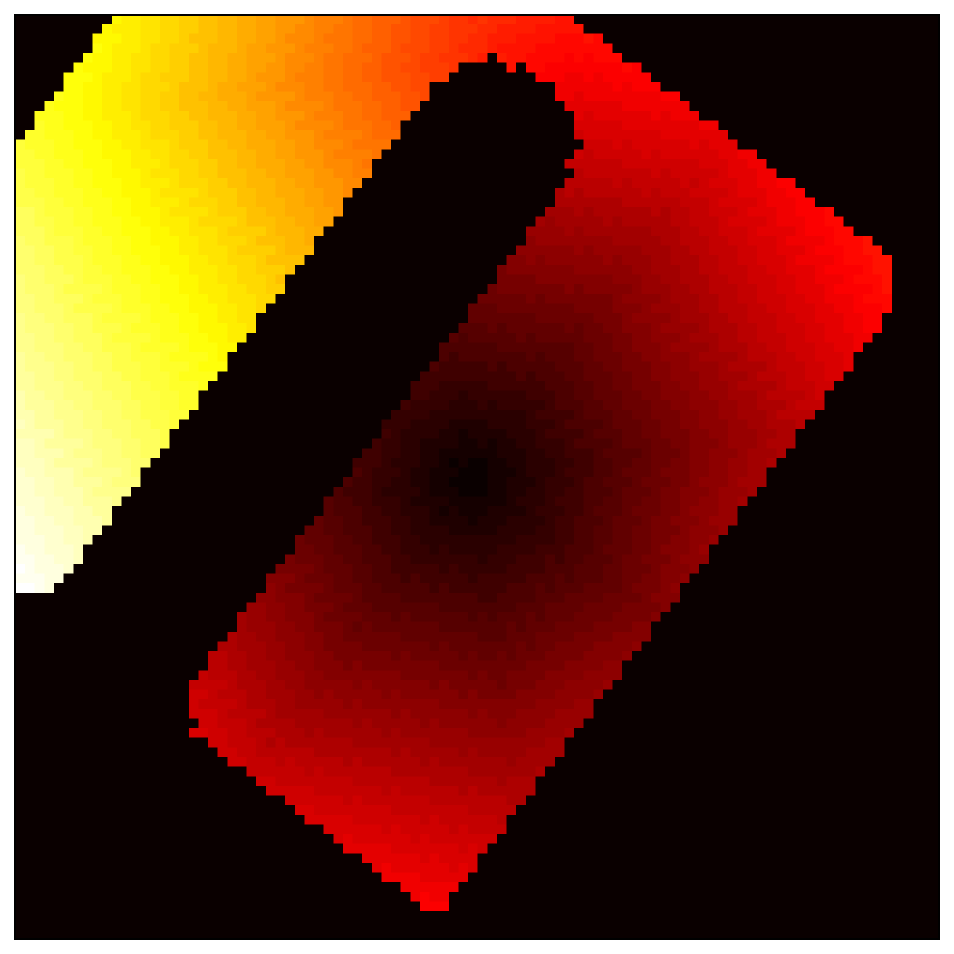}
    \caption{Shortest path from robot}
    \label{fig:obs_2}
  \end{subfigure}
  \begin{subfigure}[t]{0.24\columnwidth}
    \centering
    \includegraphics[width=\linewidth]{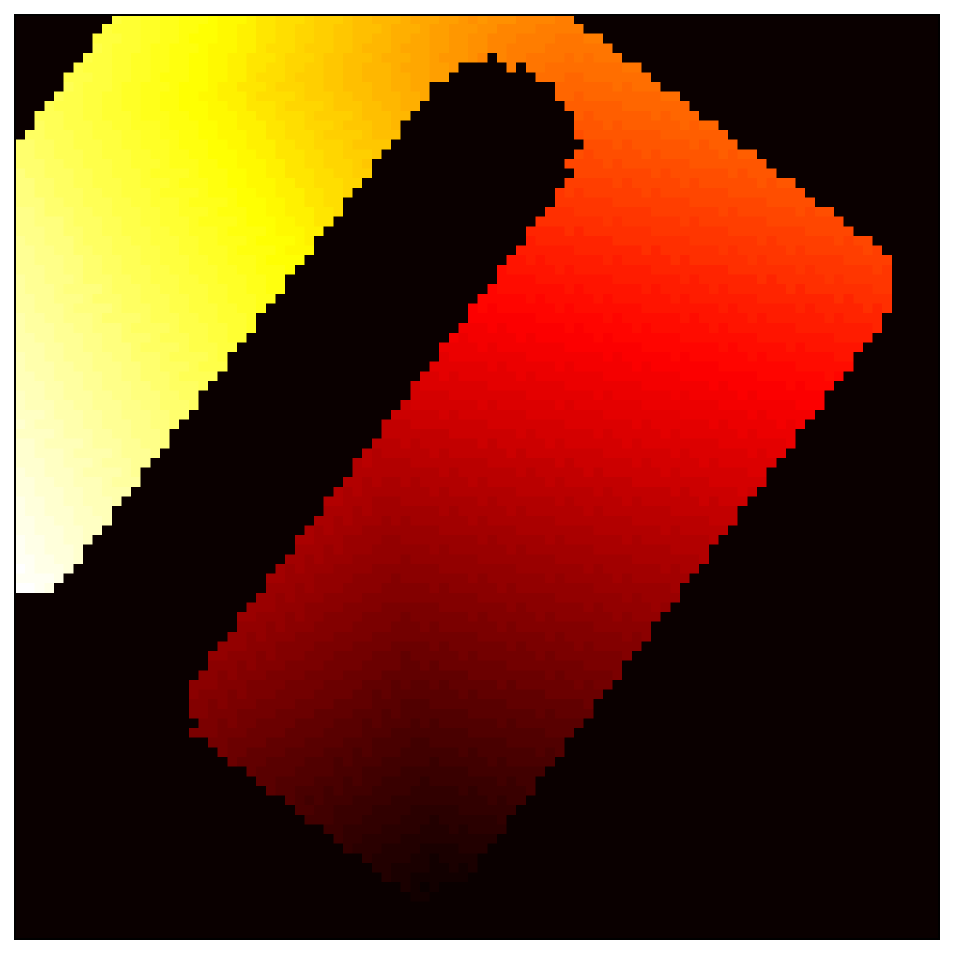}
    \caption{Shortest path to receptacle}
    \label{fig:obs_3}
  \end{subfigure}



  \caption{Visualization of the four-channel state representation in the \textit{LargeDivider} environment (colors shown for clarity only).}
\label{fig:sam_state}
\end{figure}

\noindent{\textbf{State Space.}} Each state $s \in \mathcal{S}$ is encoded as a 4-channel image aligned to the robot’s local coordinate frame, with spatial resolution $(H, W)$. The channels include: (1) A semantic segmentation map encoding object classes (e.g., obstacles, floor, boxes, robot, receptacle); (2) A binary mask representing the robot’s footprint and location; (3) A shortest-path distance map from the robot’s current position to each pixel; and (4) A shortest-path distance map from each pixel to the receptacle.  This yields a structured state space
    $\mathcal{S} = [0,1]^{H \times W \times 4}$.
An example visualization is shown in Fig. \ref{fig:sam_state}.

\vspace{0.5em}
\noindent{\textbf{Action Space.}}
Each action $a \in \mathcal{A}$ represents a pixel location $(u, v)$ in the spatial map:
    $\mathcal{A} =\{1, \dots, W\} \times \{1, \dots, H\}$.
The selected coordinate becomes the high-level goal, and the robot uses a low-level controller to navigate to it.

\vspace{0.5em}
\noindent{\textbf{Reward Function.}}
The total reward at a given step $k$ is composed of three components:
\begin{equation}
    r(k) = r_{\text{goal}}(k) + r_{\text{progress}}(k) + r_{\text{penalty}}(k),
\end{equation}
where:
\begin{alignat}{3}
&r_{\text{goal}}(k)     &=\;& +1      \; \text{for each box placed in receptacle}, \\
&r_{\text{progress}}(k) &=\;& \alpha \sum_j \Delta d_{b_j}(k), \label{eqn:cum_rew} \\
&r_{\text{penalty}}(k)  &=\;& -0.25   \; \text{for a collision or nonmovement}.
\end{alignat}

Here, $\Delta d_{b_j}(k)$ denotes the signed change in shortest-path distance from box $b_j$ to the receptacle at step $k$, and $\alpha$ is a scaling coefficient that ensures the progress reward is on a comparable scale to the other reward terms. The value of $\alpha$ is chosen empirically based on the environment size, such that progress rewards do not dominate the reward signal and overwhelm goal completion or penalty terms.
\section{Approach Overview}
\label{sec:approach}

\begin{figure*}[t]
\centering
\definecolor{uBlue}{RGB}{124,151,211}
\newcommand{\groupblock}[3][]{%
  \node[draw, rounded corners=3pt, thick, dashed,
        inner sep=6pt, fit=#2, label={[yshift=4pt]above:\textbf{#3}}, #1] {};
}
\resizebox{!}{0.22\textheight}{%
\begin{tikzpicture}[
  >=Stealth, line cap=round, node distance=6mm,
  font=\footnotesize,
  block/.style={draw, rounded corners, thick, align=center,
                minimum width=20mm, minimum height=8mm,
                top color=uBlue!30, bottom color=uBlue!30},
  map/.style={block, top color=teal!10, bottom color=teal!4},
  comp/.style={block, top color=purple!10, bottom color=purple!4},
  decision/.style={draw, diamond, aspect=2, thick, align=center, inner sep=1.5pt,
                   top color=red!8, bottom color=red!3},
  groupblock/.style={draw, rounded corners=3pt, thick, dashed,
                  inner sep=6pt, fit=#1},
  lab/.style={font=\footnotesize}
]

\node[block]    (state) {
    High-dimension state \\[0.2mm]
    \includegraphics[height=15mm]{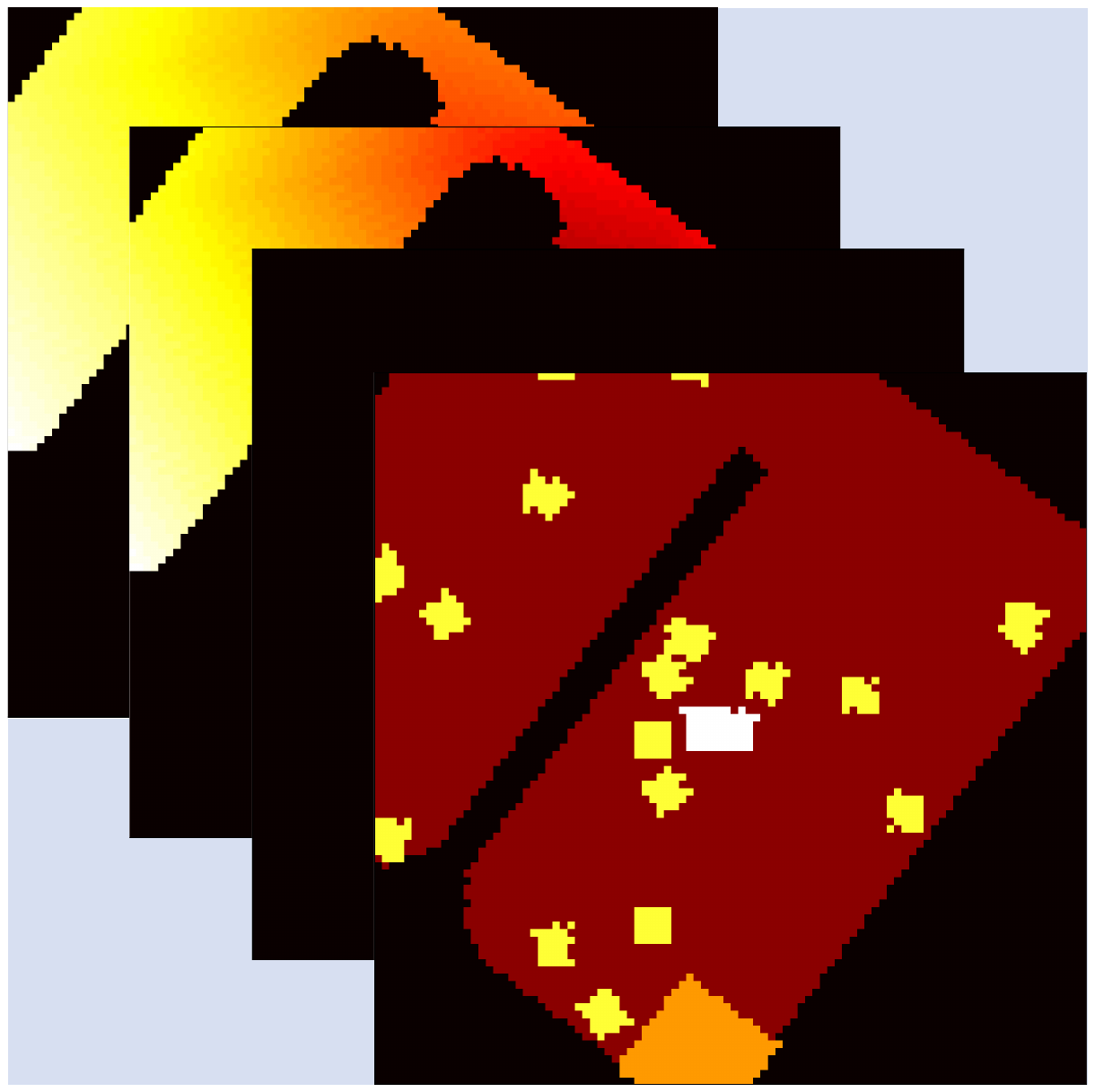}
};
\node[block] (resnet) at ($(state.south) + (0,-0.75)$) {ResNet18};
\node[map] (sam) at ($(resnet.south) + (0,-0.85)$) {Spatial Action Map};
\node[groupblock={(state)(resnet)(sam)}, label={above:\textbf{RL Policy}}] (rlbox) {};

\node[block, right=of rlbox] (spfa) {SPFA};
\node[decision, right=of spfa] (dec) {Box in path?};

\node[block] (ddpm) at ($(dec.west) + (4.75, 0)$) {Diffusion Model};
\node[block] (ldstate) at ($(ddpm.north) + (0, 0.75)$) {
    Low-dimension \\ state \\
    {\footnotesize $\left[ x_1, y_1, \ldots, x_n, y_n\right]$}
};
\node[map] (feas) at ($(ddpm.south) + (0, -0.85)$) {Feasibility Conditioning};
\node[groupblock={(ldstate)(ddpm)(feas)}, label={above:\textbf{Diffusion Policy}}] (diffbox) {};

\node[block, right=of diffbox] (ctrl) {Low-level Controller};

\coordinate (dec1) at ($(dec |- diffbox.south) + (0, -0.2)$);
\coordinate (dec2) at ($(ctrl |- diffbox.south) + (0, -0.2)$);

\draw[-Stealth, very thick] (state) -- (resnet)
                            (resnet) -- (sam);
\draw[-Stealth, very thick] (rlbox) -- (spfa);
\draw[-Stealth, very thick] (spfa) -- (dec);

\draw[-Stealth, very thick] (dec.south) -- node[left, lab]{Yes} (dec1) -- (dec2) -- (ctrl.south);
\draw[-Stealth, very thick] (dec.east) -- node[above, lab]{No} (diffbox.west);
\draw[-Stealth, very thick] (ldstate) -- (ddpm) -- (feas);
\draw[-Stealth, very thick] (diffbox.east) -- (ctrl);
\end{tikzpicture}
}
\caption{HeRD architecture. The high‑level RL policy encodes observations into a spatial action map and selects a spatial goal, which is converted to a path using SPFA. If the path intersects boxes, it is executed by a proportional controller that first rotates then translates the robot; otherwise, a diffusion policy generates a goal‑conditioned trajectory through denoising and feasibility conditioning.}
\label{fig:overview}
\end{figure*}

We design a hierarchical control framework, \textbf{HeRD}, that separates high-level semantic decision-making from low-level path planning. The high-level policy selects spatial goals using reinforcement learning with a SAM action representation \cite{wu_spatial_2020}, while a low-level controller determines how to move the robot toward that goal.

Our key contribution lies in adapting the trajectory generation strategy depending on whether the path to the spatial goal requires interacting with movable objects (i.e., boxes):

\begin{enumerate}
    \item If the SPFA-computed path to the spatial goal intersects a boxes, we retain the original path to exploit pushing behaviour.
    \item Otherwise, we generate a new trajectory using a learned diffusion policy $\pi_d$, trained on human demonstrations, which captures human-like strategies for positioning, navigation, and setup in scenarios where no immediate box interaction is required.
\end{enumerate}

Our decision to use the diffusion policy only when no boxes are intersected stems from an asymmetry in the learning signal provided by the reward function.
There is strong, immediate reward feedback when the robot pushes a box ($r_{\text{progress}}$), but when just navigating, the only available reward signal is the occasional penalty for colliding with an obstacle or not moving, and a distance-based penalty (See Eqn. \ref{eqn:sdp} in Sect. \ref{sec:RLP}). We hypothesize that the strong feedback when pushing enables the policy to specialize in this task. In contrast, there are complex and subtle concepts involved in efficiently navigating the robot that are difficult to encode into a reward function. 
Strategically repositioning or avoiding boxes en route to a spatial goal are tasks that are easy for a human to carry out, but very hard to encode into a reward function. By deferring to the diffusion policy in these situations, we leverage the intuition of the human demonstrator to generate effective trajectories for the robot.

The overall architecture is visualized in Fig.~\ref{fig:overview}.
Further details on the policy and training are provided in Section~\ref{sec:implementation}.





\section{Implementation Details}
\label{sec:implementation}
\subsection{Reinforcement Learning Policy}
\label{sec:RLP}

We formulate the high-level RL policy as a DDQN with a SAM action representation, following the architecture described in Section~\ref{sec:sam_details}, with two notable modifications to the reward function.
Qualitatively, we observe that the current policy often engages in inefficient behaviour: rather than directly pushing boxes to the receptacle, the robot tends to gather them into a corner before attempting group pushes along walls. It frequently performs backward movements or pushes boxes away from the receptacle in hopes of later combining them. This strategy leads to excessive travel and coordination failures, particularly in larger or more constrained environments.
To address this behaviour we (1) modify the $r_{\text{progress}}$ reward term and (2) add a new $r_{\text{motion}}$ term.

\vspace{0.5em}
\noindent\textbf{Progress-based reward.}
In the original SAM formulation, the robot received a reward based on the total signed progress of all boxes toward the receptacle (Eqn.~\ref{eqn:cum_rew}), 
which heavily incentivized the agent to push multiple boxes at once. While this inherently is not a negative quality, it often leads to unnecessarily long and inefficient paths in order to gather boxes to push them together.

To align with our objective of minimizing time to task completion,
we redefine the reward to consider only the signed progress of the single most-advancing box per step:
\begin{equation}
\label{eqn:max_rew} r_{\text{progress}}(k) = \alpha \cdot \Delta d_{b_j^*}(k), \quad \text{where} \; j^* = \arg \max_j |\Delta d_{b_j}(k)|,
\end{equation}
and $\Delta d_{b_j}(k)$ represents the distance box $j$ travels towards the receptacle at step $k$.
This preserves the directional component of progress while avoiding reward inflation from simultaneous multi-box pushes. 
As a result, the agent tends to act more
efficiently,
choosing to push nearby boxes individually rather than engaging in complex, inefficient maneuvers to gather and push multiple boxes at once. Nevertheless, the agent can still exploit favourable multi-box pushes if they arise naturally.

We use $\alpha = 0.2$, consistent with \cite{wu_spatial_2020}, to normalize progress rewards and prevent them from dominating other terms.

\vspace{0.5em}
\noindent\textbf{Distance-based penalty.}
To further encourage efficient trajectories, we introduce a penalty based on the robot’s displacement at each step.
The penalty is scaled to be smaller than the reward for a correct box push, ensuring that the agent
is not heavily penalized for non-pushing actions, giving it the 
\begin{equation}
\label{eqn:sdp}
r_{\text{motion}}(k) = -\frac{\alpha}{\beta} \cdot \Delta d_{\mathcal{R}}(k).
\end{equation}
Here, $\Delta d_{\mathcal{R}}(k)$ denotes the distance the robot travels at step $k$. We use $\beta=8$, but found that values in $[2,16]$ work with similar effectiveness.  

\vspace{0.5em}
\noindent{\textbf{Training Details.}}
We adopt the hyperparameters used in \cite{wu_spatial_2020}, summarized as follows: 
\emph{Replay Buffer} of 10,000 transitions; 
\emph{Optimizer} is SGD with learning rate 0.01, momentum 0.9, and weight decay 0.0001;
\emph{Loss Function} is Smooth L1 loss with gradients clipped at 10;
\emph{Batch Size} is 32; 
\emph{Discount Factor} is $\gamma = 0.99^{0.25 \cdot d_{\mathcal{R}(t)}}$, to reflect distance travelled per step (see \cite{wu_spatial_2020} for more details);
\emph{Exploration} via $\epsilon$-greedy with $\epsilon$ linearly annealed from 1.0 to 0.01 over 6,000 steps;
\emph{Training Duration} is 60,000 steps;
\emph{Random Warm-up} is 1,000 random exploration steps before training begins.

While \cite{wu_spatial_2020} trains four specialized policies, one per environment type, due to our more efficient reward design we are able to train  one generalized policy with minimal performance loss due to the generalization.
To do this, we randomly select between the \textit{LargeColumns} or \textit{LargeDivider} environments every episode (see Fig. \ref{fig:env} for the environment types), which forces the policy to adapt to various obstacle configurations. We define an episode as terminated when either all of the boxes are pushed into the receptacle or if the robot has not pushed any boxes into the receptacle for 100 steps.
We assume access to ground-truth state observations in simulation to eliminate partial observability and better isolate the impact of our reward and architecture modifications.
Training takes about 8 hours on an NVIDIA L40S GPU.

{
\setlength{\belowcaptionskip}{0pt}
\begin{figure}[htbp]
  \centering
  \begin{subfigure}[t]{0.35\columnwidth}
    \centering
    \includegraphics[width=\linewidth]{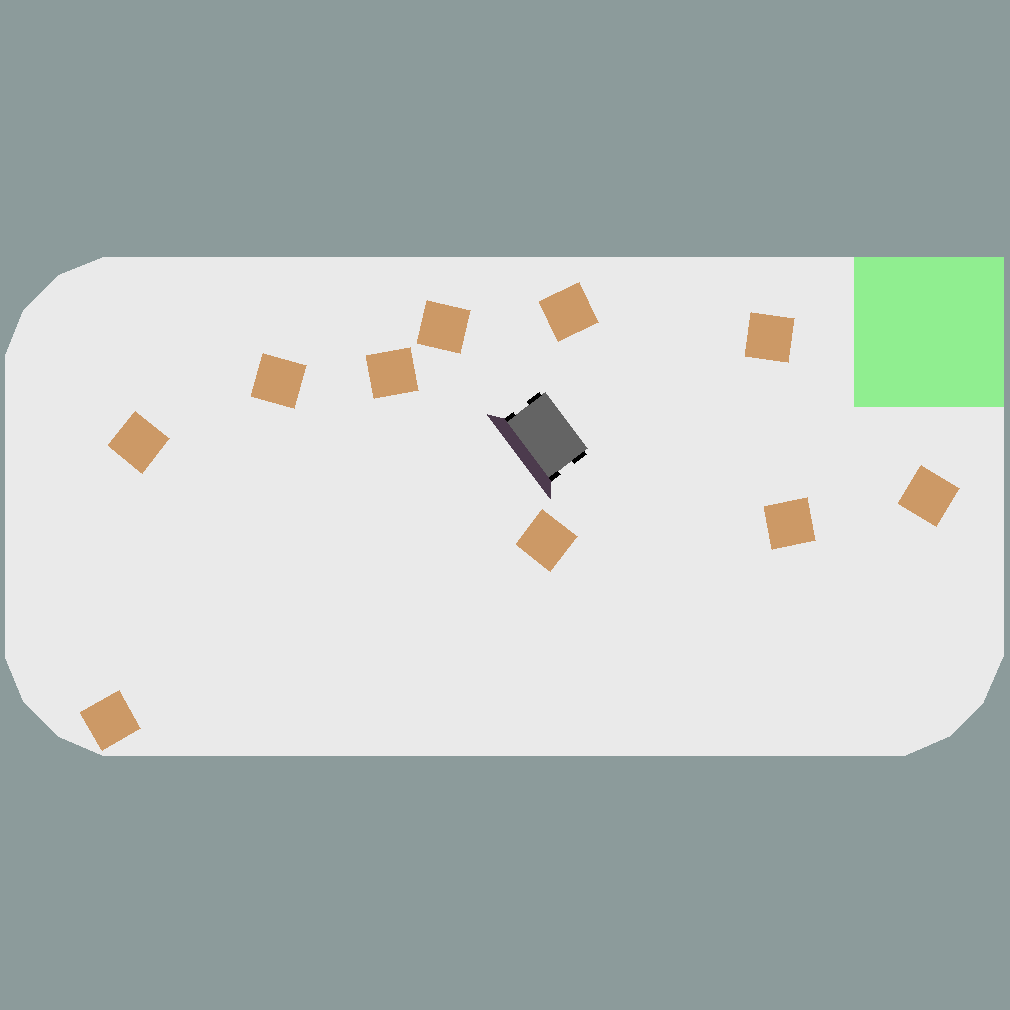}
    \caption{\textit{SmallEmpty}}
    \label{fig:a}
  \end{subfigure}\qquad
  \begin{subfigure}[t]{0.35\columnwidth}
    \centering
    \includegraphics[width=\linewidth]{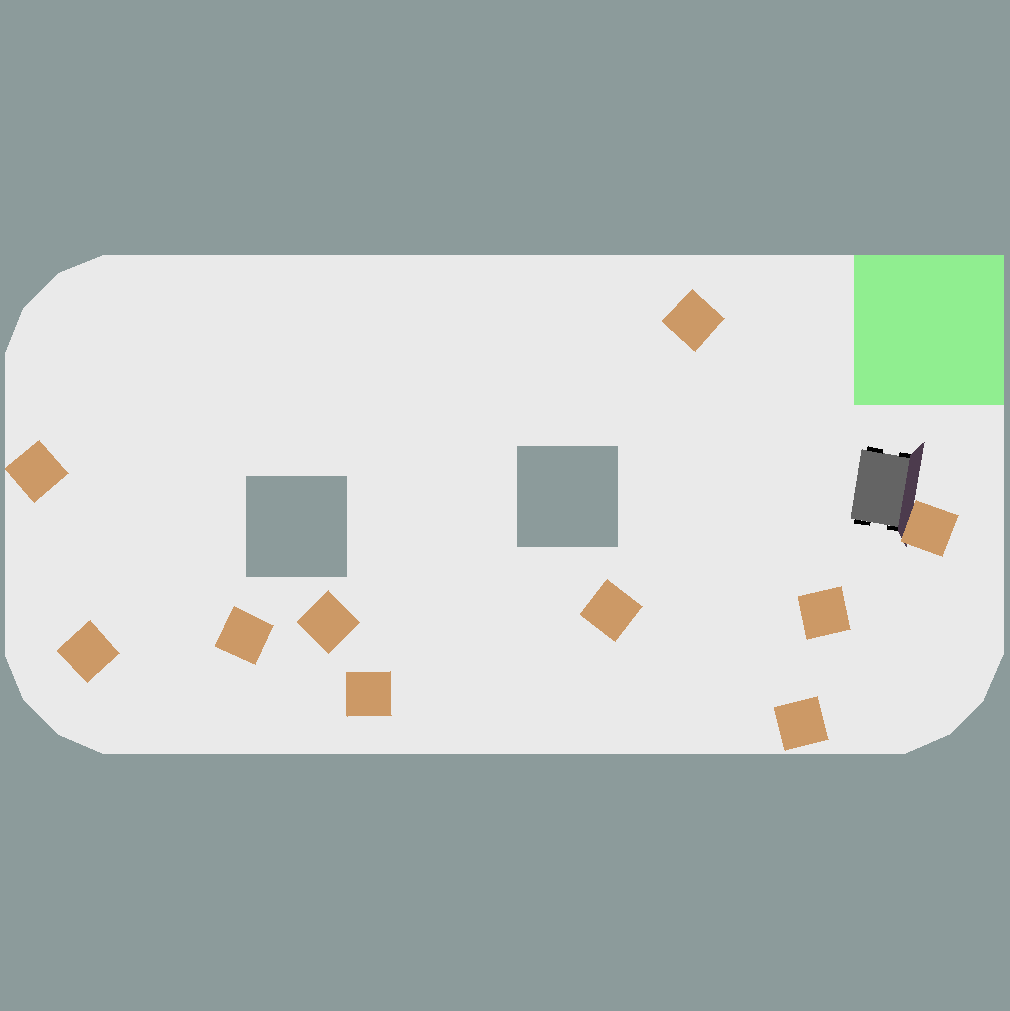}
    \caption{\textit{SmallColumns}}
    \label{fig:b}
  \end{subfigure}

  \vspace{0.75em}

  \begin{subfigure}[t]{0.35\columnwidth}
    \centering
    \includegraphics[width=\linewidth]{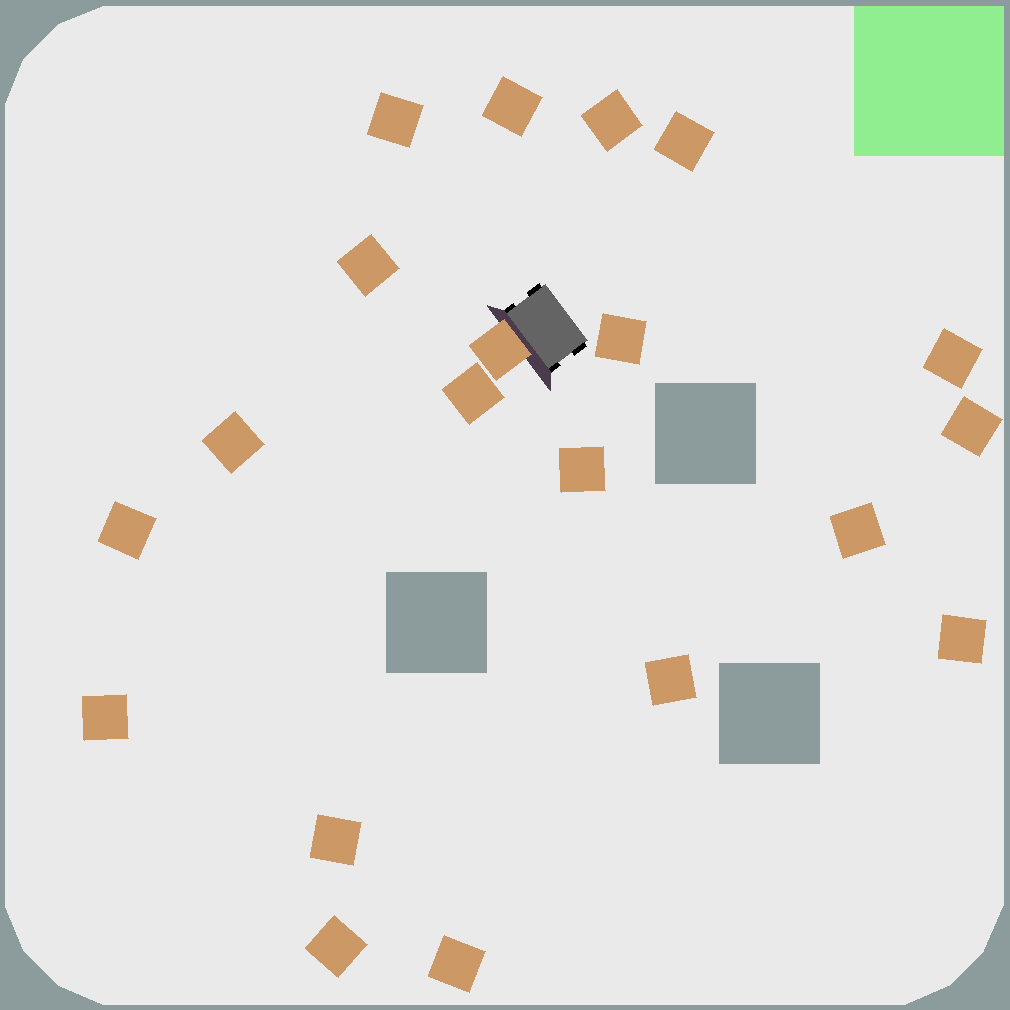}
    \caption{\textit{LargeColumns}}
    \label{fig:c}
  \end{subfigure}\qquad
  \begin{subfigure}[t]{0.35\columnwidth}
    \centering
    \includegraphics[width=\linewidth]{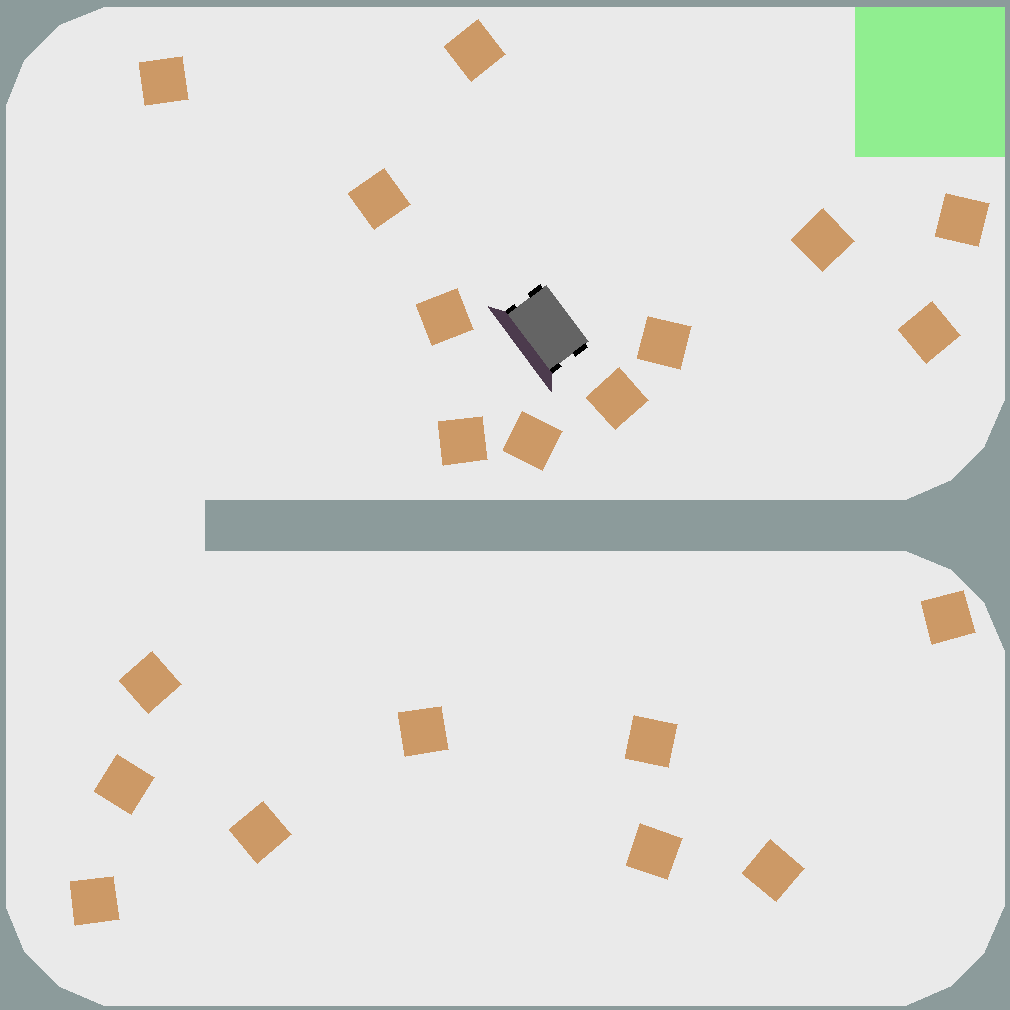}
    \caption{\textit{LargeDivider}}
    \label{fig:d}
  \end{subfigure}

  \caption{Evaluation environments used in HeRD. We use four environments varying in size and obstacle layout. The \textit{Small} environments measure 10m$\times$5m and contain 10 boxes; the \textit{Large} environments are 10m$\times$10m with 20 boxes. At each episode reset, the positions of the robot, boxes, and static obstacles are randomized. The number of columns ranges from 0–2 in the \textit{Small} environments and 0–8 in the \textit{Large} ones.}
  \label{fig:env}
\end{figure}
}

\subsection{Diffusion Policy}

\begin{figure}
\centering
    \begin{minipage}{\columnwidth}
        \centering
        \foreach \i in {2,...,5}{
            \includegraphics[width=0.36\columnwidth, angle=90]{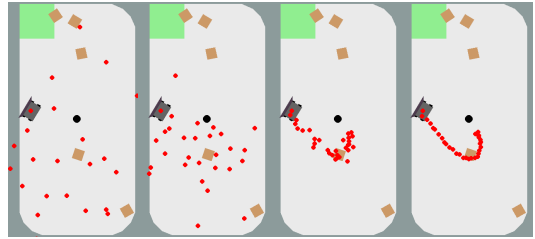} \hspace{-11pt}
        }
    \end{minipage}
\caption{Denoising process of a generated path. The high-level RL policy selects a spatial goal (the black dot in the figure), and the low-level diffusion policy generates a context-aware trajectory from the robot to the spatial goal.}
\label{fig:action}
\end{figure}

\begin{figure*}[t]
\centering
\resizebox{\textwidth}{!}{%
\begin{tikzpicture}[
    font=\sffamily,
    line cap=round,
    >=Stealth,
    node distance=1.3cm and .5cm,
    block/.style={
      draw, rounded corners, thick,
      minimum width=15mm, minimum height=15mm,
      align=center,
      top color=#1!30, bottom color=#1!10, very thick
    },
    panel/.style={ 
      draw, rounded corners, thick,
      minimum width=15mm, minimum height=45mm,
      align=center,
      top color=#1!30, bottom color=#1!30, very thick
    },
    decision/.style={ 
      draw, diamond, thick, aspect=2.2, align=center, inner sep=2pt,
      top color=red!8, bottom color=red!3, very thick
    },
    ann/.style={font=\normalsize, scale=1.6}
]
\definecolor{uBlue}{RGB}{124,151,211}
\definecolor{yCream}{RGB}{255,243,191}
\definecolor{gTeal}{RGB}{114,171,156}
\definecolor{fOrange}{RGB}{242,202,143}

\node[panel=yCream] (A) {
  \large\textbf{Goal} \\ \large\textbf{Conditioning}\\[1mm]
  \includegraphics[height=30mm]{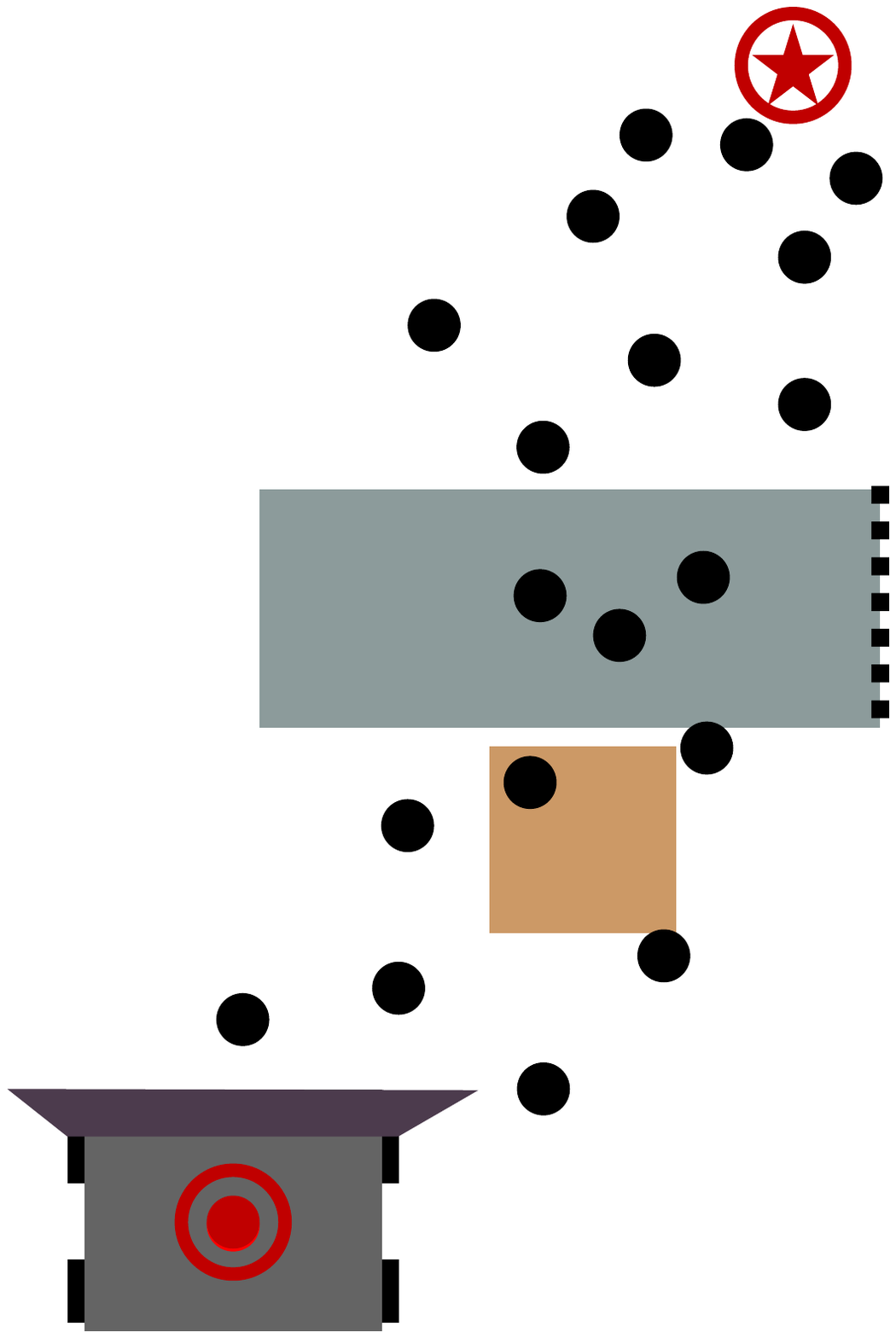}
};

\node[panel=uBlue, right=of A] (D) {
  \large\bfseries U-Net\\[2mm]
  \includegraphics[height=30mm]{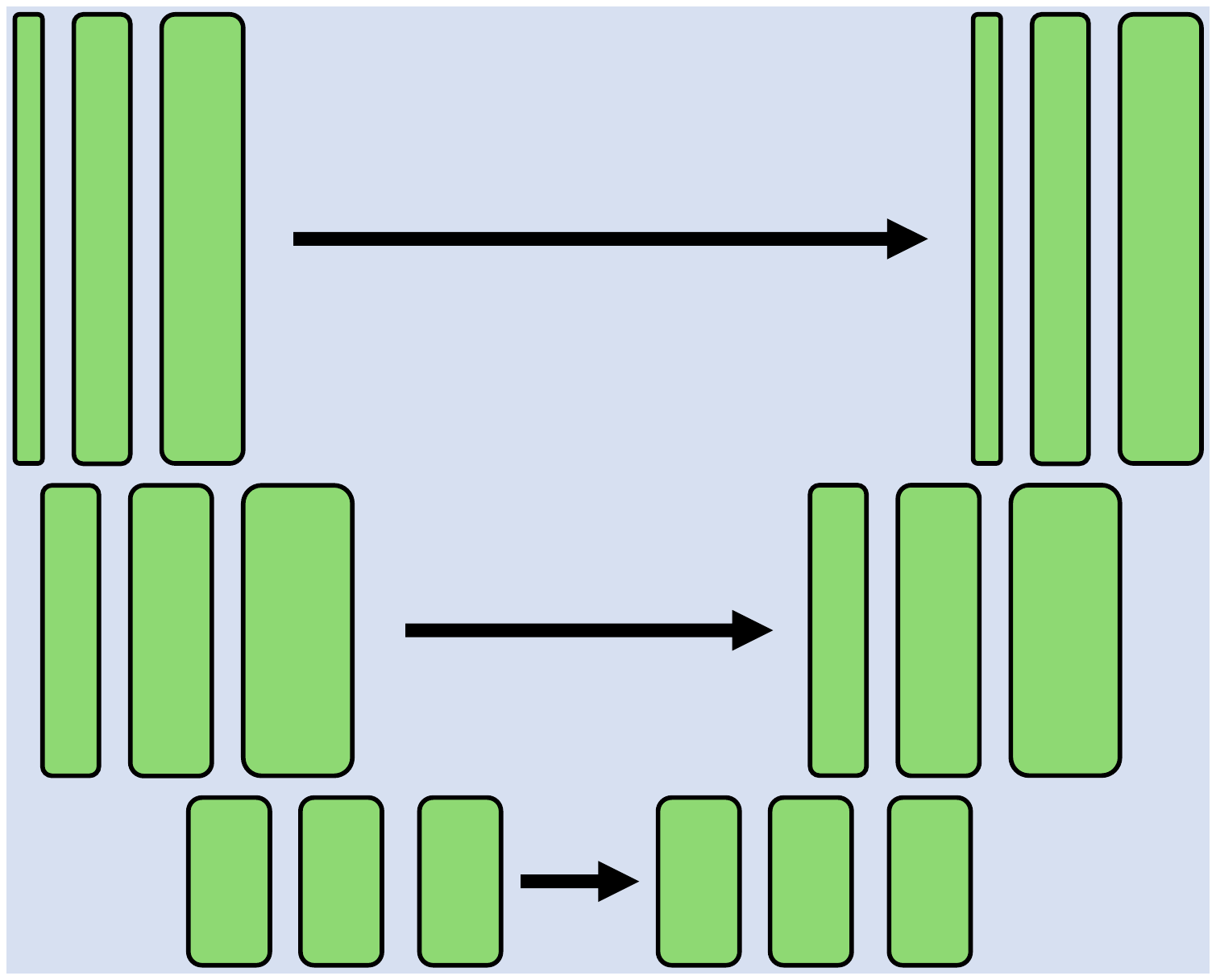}
};

\node[decision, right=of D] (Dec) {$\mathbf{i>0}$};

\node[panel=yCream, right=of Dec] (E) {
  \large\textbf{Goal} \\ \large\textbf{Conditioning}\\[2mm]
  \includegraphics[height=30mm]{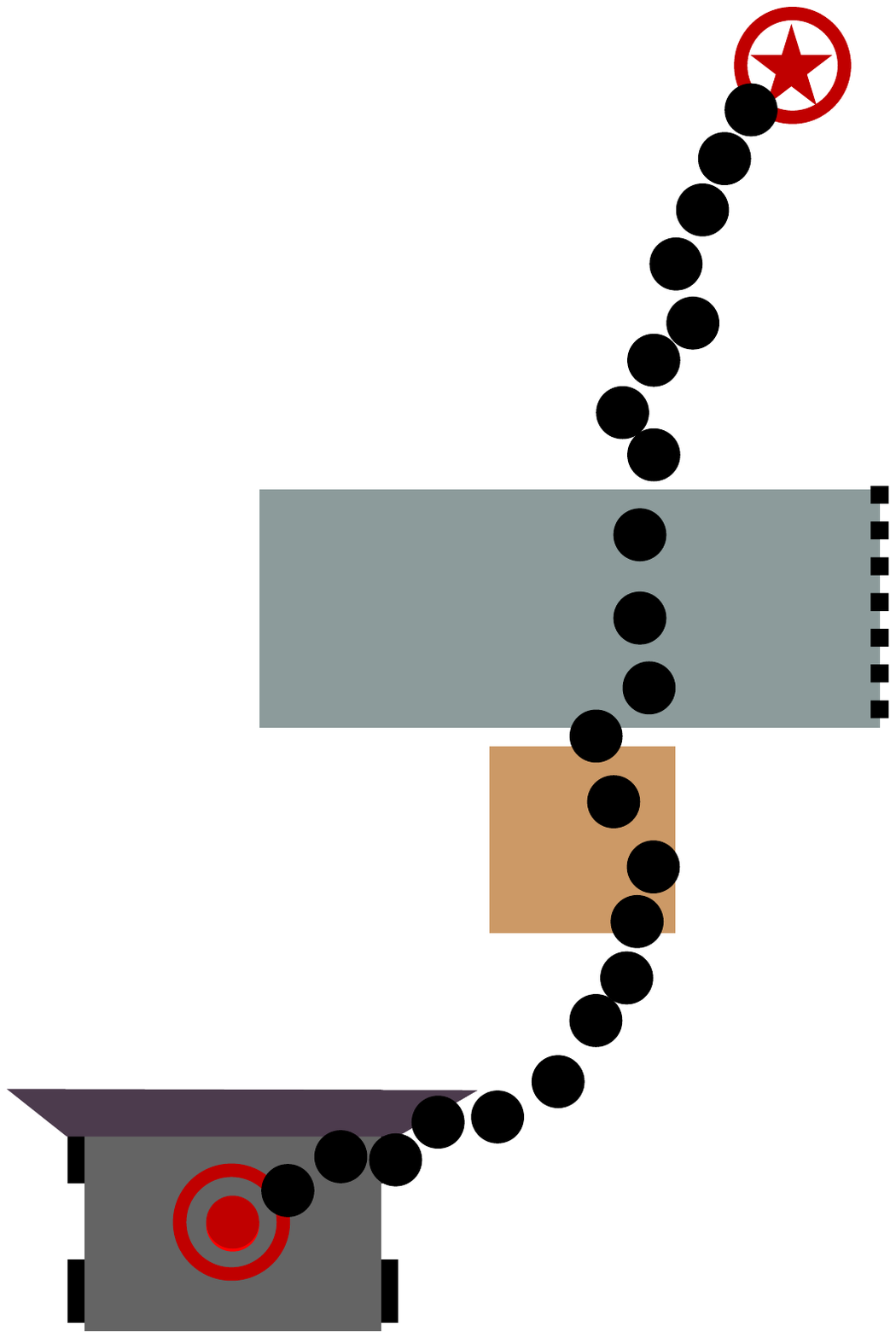}
};

\node[panel=fOrange, right=of E] (F)  {
    \large\textbf{Waypoint Feasibility}\\[2mm]
    \includegraphics[height=30mm]{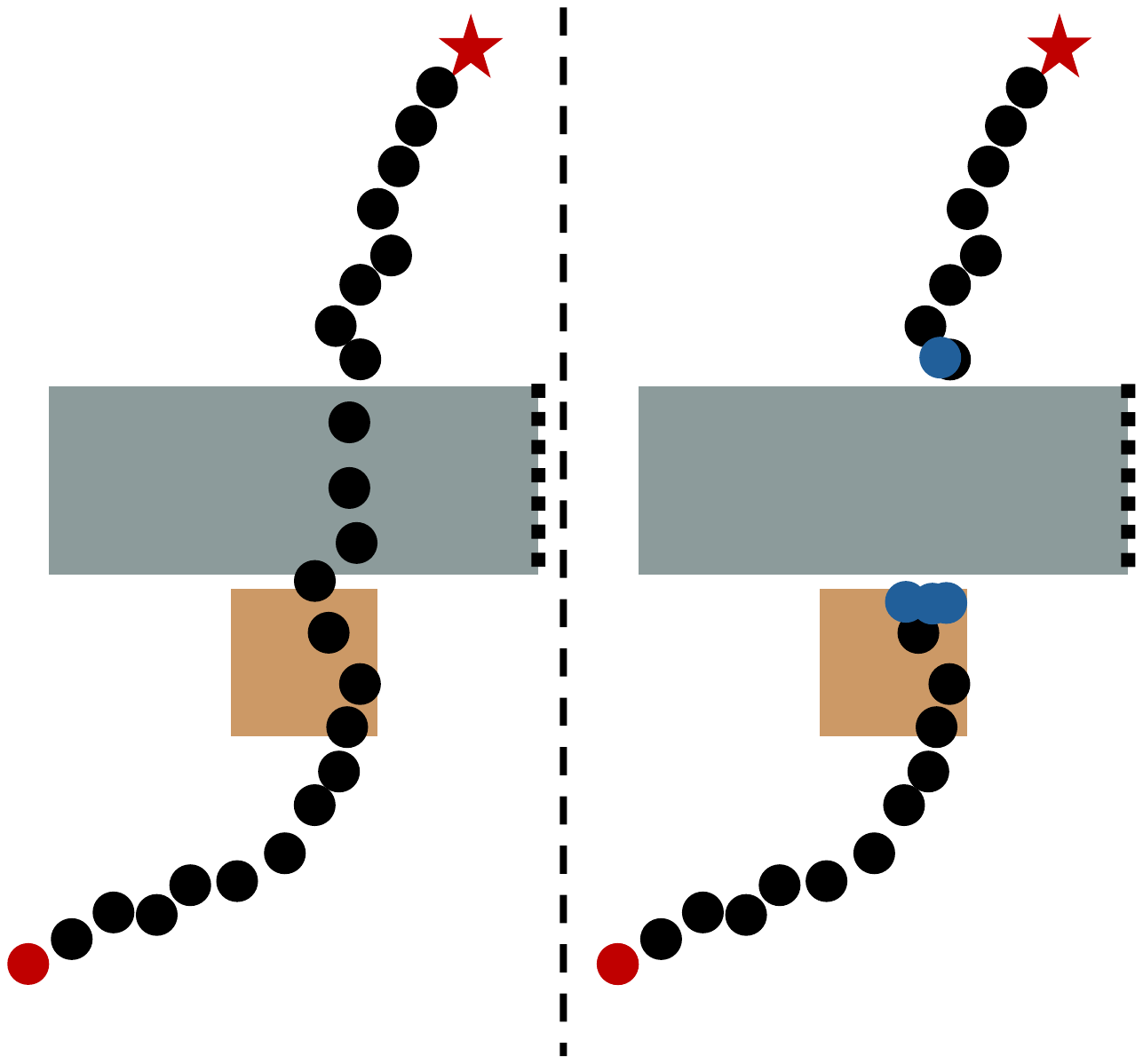}
};

\node[panel=gTeal, right=of F] (G) {
  \large\textbf{Path Pruning}\\[2mm]
  \includegraphics[height=30mm]{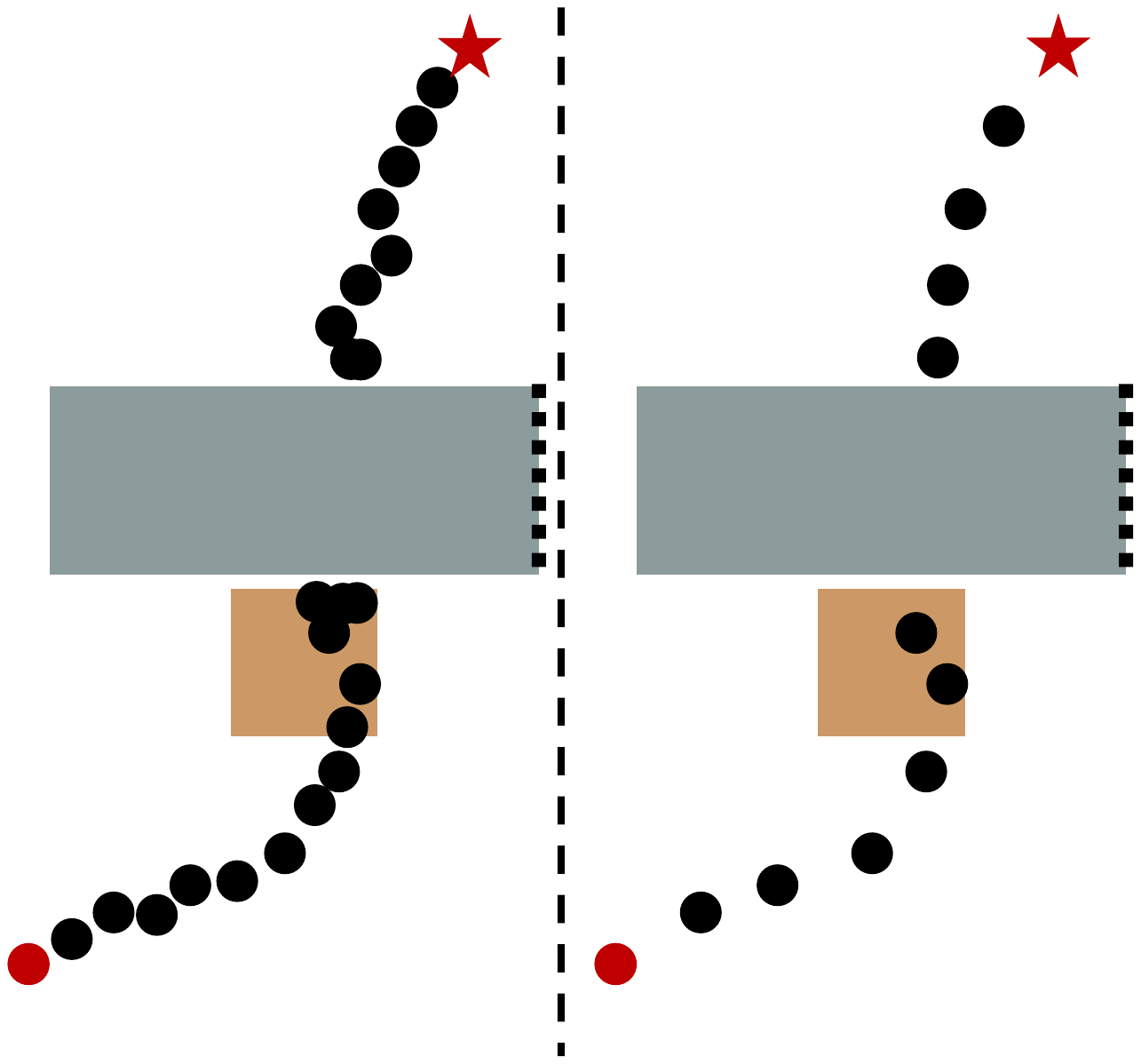}
};

\node[panel=fOrange, right=of G] (H) {
  \large\textbf{Path Feasibility}\\[2mm]
  \includegraphics[height=30mm]{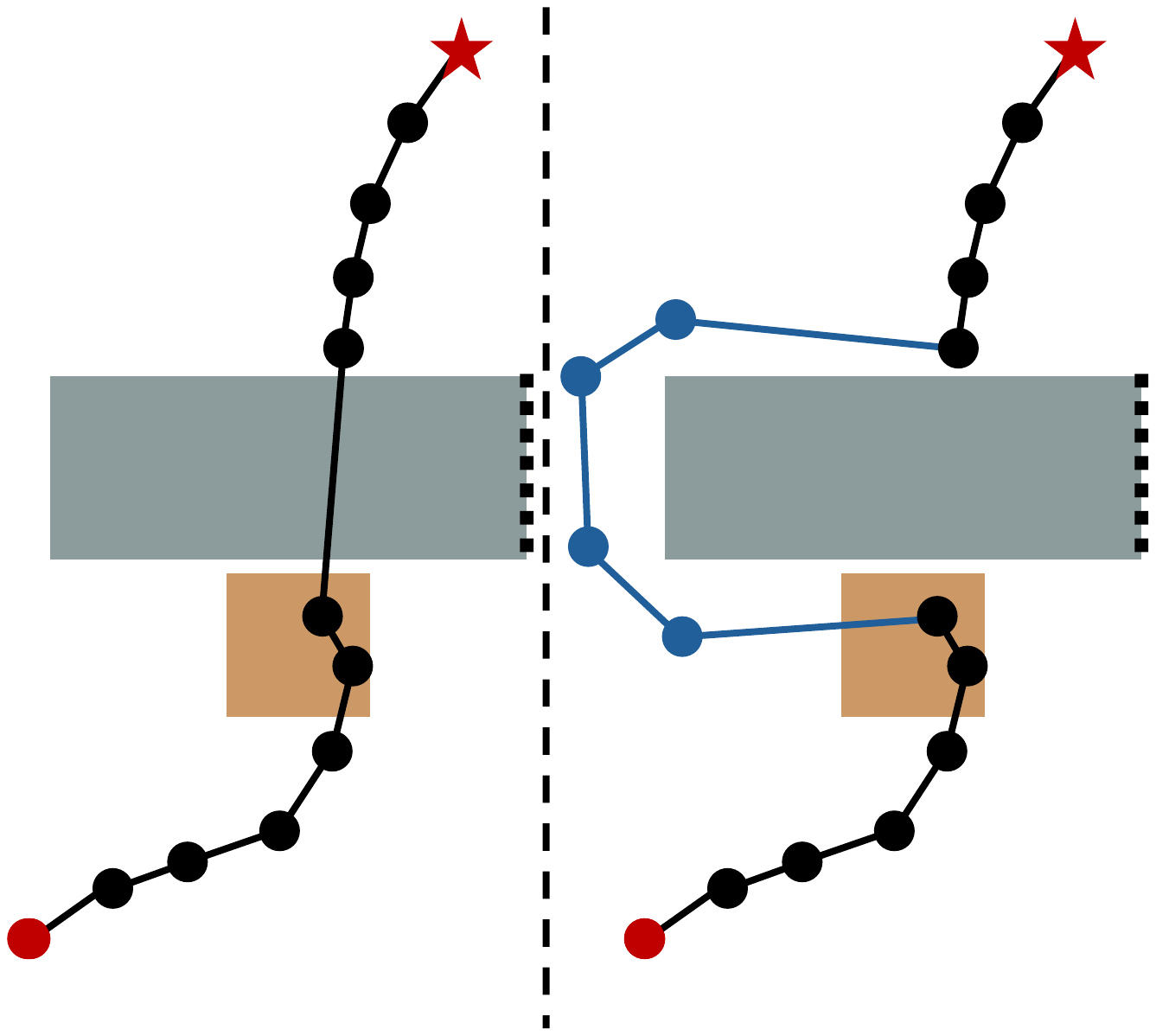}
};

\coordinate (inL) at ($(A.west)+(-.75,0)$);
\draw[-Stealth, very thick] (inL) -- (A.west);
\draw[-Stealth, very thick] (A) -- (D);
\draw[-Stealth, very thick] (D) -- (Dec);
\draw[-Stealth, very thick] (Dec) -- (E);
\draw[-Stealth, very thick] (E) -- (F);
\draw[-Stealth, very thick] (F) -- (G);
\draw[-Stealth, very thick] (G) -- (H);

\draw[-Stealth, very thick] (H.east) -- ++(0.5,0);

\coordinate (fbR) at ($(Dec |- A.south)+(0,-1.0)$);
\coordinate (fbL) at ($(A.south west)+(-0.5,-1.0)$);

\draw[-Stealth, very thick]
  (Dec.south) -- node[right, ann]{\footnotesize Yes} (fbR)
  -- (fbL) -- ($(fbL |- A)$);

\node[ann] at ($(A.west)+(-1,0.1)$) {\small$\tau^N$};
\node[ann] at ($(H.east)+(0.5,0.4)$) {\small$\tau^0$};
\node[ann] at ($(A.north)+(0,1)$) {\small$\tau^{i}_{\mathrm{term.}}\leftarrow x_{\mathrm{goal}}$};
\node[ann] at ($(A.north)+(0,0.4)$) {\small$\tau^{i}_{0}\leftarrow x_{\mathrm{robot}}$};
\node[ann] at ($ (fbL)!0.5!(fbR) + (0,-.6) $) {\small$i=N,\ldots,0$};
\node[ann] at ($ (Dec.east)!0.3!(E.west) + (0,.4) $) {\footnotesize No};

\end{tikzpicture}%
}
\caption{Diffusion-based trajectory generation pipeline. At each difusion timestep $i$, the trajectory $\tau^i$ is denoised using a 1D U-Net conditioned on observation features. The trajectory’s endpoints are fixed via goal inpainting: the first point is the robot’s position, and the last is the spatial goal. If $i > 0$, the intermediate points are further denoised. Once denoising completes ($i = 0$), the trajectory undergoes a feasibility pipeline: each waypoint is projected to free space if necessary, redundant waypoints are pruned, and path segments crossing obstacles are repaired via SPFA. The final output is a smooth trajectory that respects geometric constraints.}
\label{fig:denoise}
\end{figure*}

Our diffusion-based controller generates trajectories from demonstration data to guide the robot toward a spatial goal in non-pushing scenarios. It is queried during execution when the SAM-selected goal is reachable without box interaction. Fig.~\ref{fig:action} shows one such diffused path.

\vspace{0.5em}
\noindent{\textbf{Model Architecture and Conditioning.}}
Following~\cite{chi_diffusion_2024}, we implement a conditional diffusion model using a 1D U-Net for denoising, conditioned via FiLM layers \cite{perez_film_2018} on low-dimensional observation features.
Unlike the receding-horizon approach used in~\cite{chi_diffusion_2024}, we generate the entire trajectory in a single forward pass. This is necessary to support goal conditioning via inpainting~\cite{janner_planning_2022}, where at every denoising step we fix the first point of the trajectory to the robot's current position and the last point to the spatial goal. This guides the waypoints into a coherent trajectory. Receding-horizon rollouts are incompatible here, as they lack a consistent final waypoint.

The denoising process is defined iteratively as
\begin{equation}
    \tau_t^{i-1} = a\left( \tau_t^i - \lambda \, \epsilon_\theta(\mathbf{O}_t, \tau_t^i, i) + \mathcal{N}(0, \sigma^2 I) \right),
    \label{eqn:condDDPM}
\end{equation}
where $\tau^i$ is the noisy trajectory at timestep $i$, $\mathbf{O}_t$ is the observation vector, $\epsilon_\theta$ is the noise prediction network, $a(\cdot)$ enforces the goal conditioning, and $\lambda\in(0,1)$ is the discount factor in the diffusion process.
The model is trained to minimize the standard denoising loss: $\mathcal{L} = \text{MSE} \left( \epsilon^i, \epsilon_\theta(\mathbf{O}_t, \tau^0 + \epsilon^i, i) \right)$, and generates 32-point trajectories using DDIM sampling at inference time.

\vspace{0.5em}
\noindent{\textbf{Demonstration Data and Augmentation.}}
To collect expert demonstrations for training the diffusion policy, we deployed a pretrained high-level policy in a large, obstacle-free environment (\textit{LargeEmpty}) to generate spatial goals. For each episode, a human demonstrator teleoperated the robot to navigate toward the SAM-selected goal while strategically considering the most effective path to the goal. These considerations included grouping boxes against walls, avoiding interactions with boxes that would be easier to push later, and pushing multiple boxes at once.

Each demonstration episode begins at the robot’s initial position and ends at the SAM-specified goal location. Trajectories are recorded by logging the robot’s $(x, y)$ position at fixed intervals of 0.3 meters of travel, producing a sequence of sparse, evenly spaced waypoints. To improve smoothness and standardize input dimensions for the diffusion policy, we then apply linear interpolation between these waypoints to generate a dense 32-point trajectory per episode.

Approximately 1,000 episodes were collected. To augment the dataset, we generate additional demonstrations by selecting any
non-interpolated 
waypoint from an existing trajectory as the new starting position, and
padding the end of the trajectory with the spatial goal to maintain the fixed length.
This yields $\sim$13,500 valid demonstrations for training.

Each demonstration is paired with a state observation composed of: (1) the 4-corner vertices of the robot and receptacle, (2) the $(x, y)$ positions of the four closest boxes to the robot, and (3) the spatial goal. If fewer than four boxes remain in the environment, placeholder boxes centered within the receptacle are used to preserve a fixed input size.
All state inputs are normalized following the protocol from~\cite{chi_diffusion_2024}. We made the design decision to include only the four closest boxes based on our empirical hypothesis that human demonstrators struggled to consider more than four boxes when teleoperating the robot. In addition, we found that the orientation of the boxes was not essential to include in the state, allowing us to further reduce the state space by only including the box positions.

\vspace{0.5em}
\noindent{\textbf{Training Details}.}  The diffusion model was trained for 3,500 epochs using a batch size of 256, with a validation split of $2\%$. Optimization was performed using AdamW ($\beta_1 = 0.95$, $\beta_2 = 0.999$, $\epsilon = 10^{-8}$, weight decay $= 10^{-6}$, learning rate $= 10^{-4}$). Validation was performed every 10 epochs. Although full training took approximately 1 hour on a single NVIDIA L40S GPU, we selected the checkpoint with the lowest validation loss, which occurred at epoch 350. This corresponds to roughly 6 minutes of training time in practice.
We trained with 100 denoising steps but use 15-step DDIM sampling at inference to improve efficiency.

\vspace{0.5em}
\noindent{\textbf{Trajectory Postprocessing.}}
To ensure the diffusion path is feasible, we use the following postprocessing procedure:
\begin{enumerate}
    \item \emph{Goal Conditioning:} All sampled trajectories should start at the robot and end at the spatial goal. We use conditioning-by-inpainting to replace the first and last sampled points in the trajectory with the robot's position and spatial goal.
    
    \item \emph{Waypoint Feasibility:} Each waypoint is checked for collision; if it lies within an obstacle, it is projected to the nearest valid point in $\mathcal{W}_{free}$.

    \item \emph{Pruning:} Waypoints are sparsified via a distance threshold. This results in a more efficient trajectory without compromising fidelity.

    \item \emph{Trajectory Feasibility:} Even if all waypoints lie in free space, sequential points may straddle obstacles. Line segments crossing obstacles are repaired using SPFA.
\end{enumerate}
The complete denoising process is shown in Fig. \ref{fig:denoise}.

\section{Results}

\begin{table}
\caption{HeRD vs. Baseline Performance}
\label{tab:baseline}

\renewrobustcmd{\bfseries}{\fontseries{b}\selectfont}
\renewrobustcmd{\boldmath}{}

\centering
\resizebox{\columnwidth}{!}{%
\small
\begin{tabular}{
    l 
    S[table-format=2.2(2),detect-weight,mode=text]
    S[table-format=2.0(2),detect-weight,mode=text] 
    S[table-format=2.2(2),detect-weight,mode=text]
    S[table-format=2.0(2),detect-weight,mode=text]
}
\toprule
& \multicolumn{2}{c}{Ours}
& \multicolumn{2}{c}{\shortstack{Baseline}} \\
\cmidrule(lr){2-3} \cmidrule(lr){4-5}
Environment & {Boxes} & {Distance}
            & {Boxes} & {Distance} \\
\midrule
SmallEmpty   & \bfseries 10.00 \pm 0.00 & \bfseries 115 \pm 23
             & 9.90 \pm 0.30            & 219 \pm 76 \\
SmallColumns & \bfseries 10.00 \pm 0.00 & \bfseries 114 \pm 31
             & 9.60 \pm 0.80            & 271 \pm 110 \\
LargeColumns & \bfseries 19.80 \pm 0.51  & \bfseries 339 \pm 67 
             & 18.70 \pm 3.08           & 551 \pm 206 \\
LargeDivider & \bfseries 19.6 \pm 0.92 & \bfseries 345 \pm 69
             & 18.85 \pm 1.39           & 705 \pm 212 \\
\bottomrule
\end{tabular}%
}
\end{table}
\begin{table*}[ht]
\caption{Effect of the different components of HeRD}
\label{tab:ablation}

\renewrobustcmd{\bfseries}{\fontseries{b}\selectfont}
\renewrobustcmd{\boldmath}{}

\centering
\resizebox{\textwidth}{!}{%
\small
\begin{tabular}{
    l 
    S[table-format=2.2(2),detect-weight,mode=text]
    S[table-format=2.0(2),detect-weight,mode=text] 
    S[table-format=2.2(2),detect-weight,mode=text]
    S[table-format=2.0(2),detect-weight,mode=text] 
    S[table-format=2.2(2),detect-weight,mode=text]
    S[table-format=2.0(2),detect-weight,mode=text]
    S[table-format=2.2(2),detect-weight,mode=text]
    S[table-format=2.0(2),detect-weight,mode=text]
    S[table-format=2.2(2),detect-weight,mode=text]
    S[table-format=2.0(2),detect-weight,mode=text]
}
\toprule
& \multicolumn{2}{c}{Ours}
& \multicolumn{2}{c}{\shortstack{No diffusion policy}}
& \multicolumn{2}{c}{\shortstack{Only diffusion policy}}
& \multicolumn{2}{c}{\shortstack{Cumulative\\progress reward}}
& \multicolumn{2}{c}{\shortstack{No step\\distance penalty}} \\
\cmidrule(lr){2-3} \cmidrule(lr){4-5} \cmidrule(lr){6-7} \cmidrule(lr){8-9} \cmidrule(lr){10-11}
Environment & {Boxes} & {Distance}
            & {Boxes} & {Distance}
            & {Boxes} & {Distance}
            & {Boxes} & {Distance}
            & {Boxes} & {Distance} \\
\midrule
SmallEmpty   & \bfseries 10.00 \pm 0.00 & \bfseries 115 \pm 23
             & 9.95 \pm 0.22            & 130 \pm 38
             & 9.55 \pm 0.5             & 246 \pm 79
             & \bfseries 10.00 \pm 0.00 & 150 \pm 39
             & \bfseries 10.00 \pm 0.00 & 119 \pm 25 \\
SmallColumns & \bfseries 10.00 \pm 0.00 & 114 \pm 31
             & \bfseries 10.00 \pm 0.00 & 131 \pm 25
             & 9.65 \pm 0.48            & 200 \pm 90
             & 9.80 \pm 0.87            & 138 \pm 42
             & \bfseries 10.00 \pm 0.00 & \bfseries 104 \pm 26 \\
LargeColumns & \bfseries 19.80 \pm 0.51 & \bfseries 339 \pm 67 
             & 19.55 \pm 0.86           & 403 \pm 136
             & 18.90 \pm 1.22           & 540 \pm 186
             & 18.75 \pm 2.81           & 435 \pm 106
             & 19.15 \pm 3.26           & 360 \pm 77 \\
LargeDivider & \bfseries 19.6 \pm 0.92 & 345 \pm 69
             & 19.90 \pm 0.44           & 547 \pm 138
             & 19.15 \pm 0.85           & 639 \pm 141
             & 18.90 \pm 1.41           & 408 \pm 95
             & 19.40 \pm 1.24           & \bfseries 318 \pm 58 \\
\bottomrule
\end{tabular}%
}
\end{table*}

We evaluate the performance of HeRD on the \textit{Box-Delivery} task from Bench-NPIN \cite{zhong_bench-npin_2025}, where the objective is to push scattered boxes into a receptacle while navigating through randomized environments with varying levels of fixed obstacles. We test across four environments of increasing difficulty, as illustrated in Fig. \ref{fig:env}. Each trial initializes the robot, boxes, and obstacles at random positions. For fair comparison, all models are evaluated on the same random seed, with 20 trials per environment.

We report two primary metrics: (1) the number of boxes successfully delivered to the receptacle, and (2) the total distance travelled by the robot within the episode.

\vspace{0.5em}
\noindent \textbf{Comparison to baseline.}
We first compare HeRD to the state-of-the-art spatial action maps (SAM) formulation presented in \cite{wu_spatial_2020}. As in the original implementation, we train a separate SAM model per environment. In contrast,
we train a single generalized HeRD policy.

\begin{figure}[tbp]
  \centering
    \begin{subfigure}[t]{0.47\columnwidth}
    \centering
    \includegraphics[width=\linewidth]{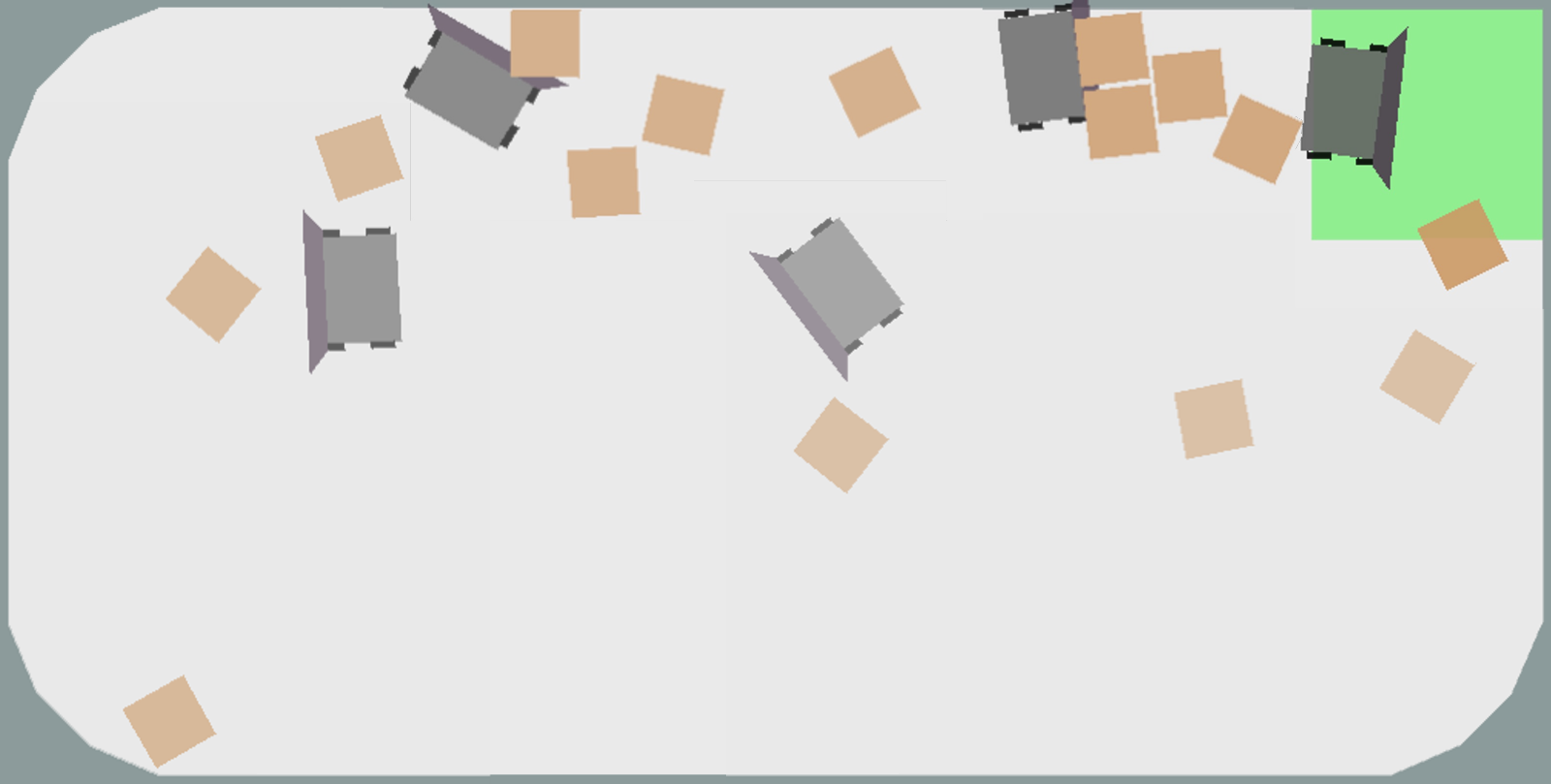}
    \caption{HeRD}
    \label{fig:ours}
  \end{subfigure} \quad
  \begin{subfigure}[t]{0.47\columnwidth}
    \centering
    \includegraphics[width=\linewidth]{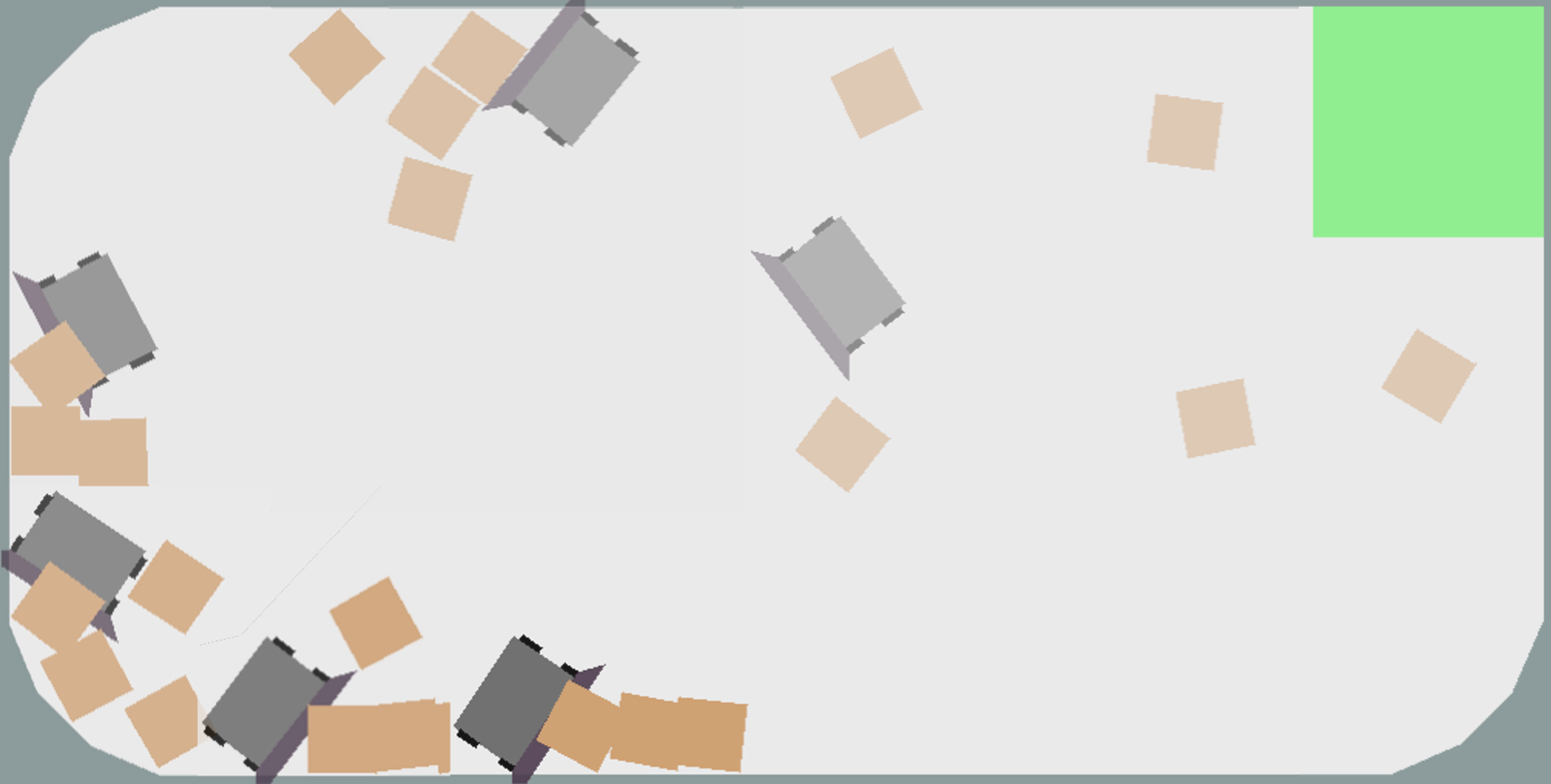}
    \caption{Baseline}
    \label{fig:base}
  \end{subfigure}

  \caption{The difference in behaviour of HeRD compared to the baseline. The figure shows the first few actions of each policy in the same environment. HeRD immediately seeks out boxes to push into the receptacle while the baseline focuses on gathering multiple boxes together, even pushing them backwards to do so.}
\label{fig:behaviour}
\end{figure}

Evaluation results in Table~\ref{tab:baseline} show that HeRD achieves both higher success rates and significantly shorter paths than the SAM baseline.
As we discuss in further detail below, the reward modifications improve the behaviour of HeRD by deterring the inefficient box-grouping actions seen in the baseline, shown in Fig.~\ref{fig:behaviour}. This combined with the generative paths account for the increased success of HeRD. 
For instance, in \textit{LargeDivider}, the baseline travels an average of $705$ meters, representing a $104\%$ increase in distance travelled compared to HeRD. Despite being trained as a single generalized policy, HeRD consistently outperforms these specialized baselines in both efficiency and success.

\vspace{0.5em}
\noindent \textbf{Effect of separating pushing and non-pushing actions.}
We hypothesize that the high-level policy specializes in pushing actions but struggles to plan efficient navigation actions,
which is why we utilize diffusion for these scenarios.  To validate this design, we compare HeRD to two ablations: one trained \textit{without} a diffusion policy, and one that uses the diffusion policy \textit{exclusively} for all trajectories. See Table~\ref{tab:ablation}.

The “No diffusion trajectories” variant outperforms the “Only diffusion trajectories” variant in both success rate and efficiency. This supports our claim that the high-level policy excels at pushing, and also highlights the limitations of relying entirely on diffusion-generated paths. Diffusion trajectories lack the fine-grained precision required for effective pushing, especially in cluttered or multi-box scenarios. A much larger demonstration dataset would be required to overcome this.

\begin{figure}
    \centering
    \resizebox{\columnwidth}{!}{%
    \includegraphics{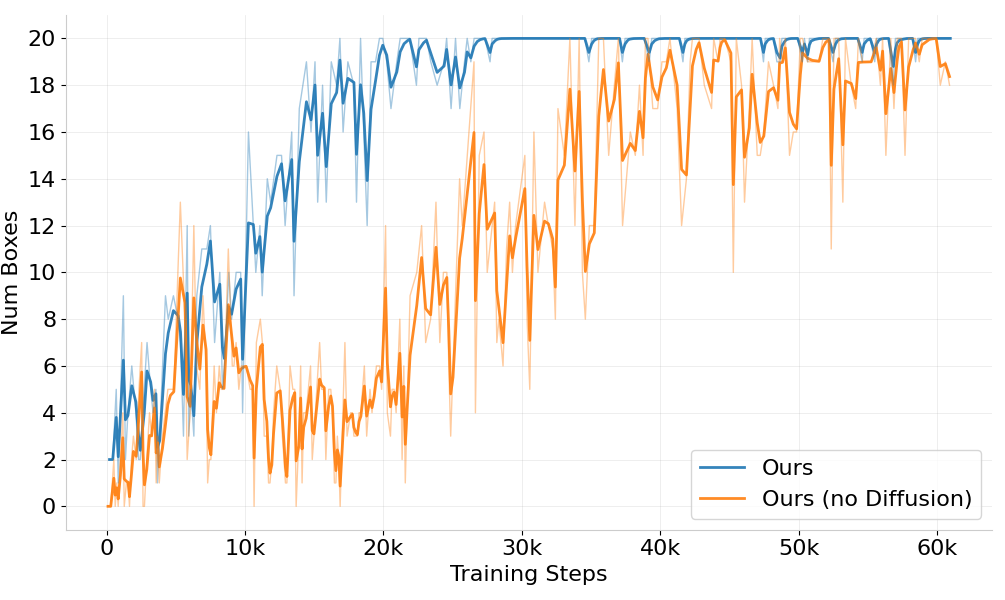}
    }
    \caption{Training curves for the high-level policy. HeRD converges substantially earlier than the no-diffusion variant, reflecting improved learning efficiency.}
    \label{fig:training_curve}
\end{figure}

HeRD outperforms both ablations, indicating that combining high-level policy competence with selectively deployed diffusion trajectories offers the best of both worlds. We observe improved sample efficiency as well: Fig. \ref{fig:training_curve} shows that HeRD converges roughly 16,000 samples earlier than the no-diffusion variant. This suggests that delegating low-reward decision-making to the diffusion model helps smooth the learning signal and accelerates training.

\vspace{0.5em}
\noindent \textbf{Effect of reward structure.} 
We perform two ablations to evaluate how the reward function components contribute to performance. The first reverts our maximum-based progress reward (Eqn. \ref{eqn:max_rew}) back to the cumulative formulation (Eqn. \ref{eqn:cum_rew}). The second removes the step distance penalty (Eqn. \ref{eqn:sdp}) to assess its impact on efficiency and task success.

The results under the “Cumulative progress reward” column in Table~\ref{tab:ablation} show a noticeable decline in efficiency and, in some environments, task success. This formulation encourages the robot to gather multiple boxes before pushing to maximize per-step reward. While this behaviour can be effective in isolated cases, it often leads to complex setup actions, including backward movement and detours. In constrained environments like \textit{LargeDivider}, this results in longer paths and reduced robustness. By contrast, our maximum-based formulation reduces this bias and encourages steady, directed progress toward the goal, even when pushing a single box. This leads to more robust and efficient behaviour.

The results of the “No step distance penalty” experiment yield the most competitive metrics compared to ours. However, removing the penalty causes the success rate to suffer in the more complex environments. While the changes are subtle, the penalty appears to help discourage inefficient motion while retaining exploration. 

\subsection{Real-World Experiments}
We evaluate HeRD and a baseline model in a real-world setup to test the effectiveness of the policies in a physical environment. For the experiments, we used a TurtleBot3 Burger with a 3D printed bumper fixed to the front, and recreated the \textit{SmallEmpty} environment in a lab. Using an overhead camera and visual markers, we obtain accurate pose estimations of the robot, boxes, receptacle, and environment boundaries, which we then use to construct the observations for the robot. Our setup is shown in Fig~\ref{fig:physical}. From this, we test the policies trained in simulation. Given the generated observations, the policy generates a spatial goal that is then carried out on the robot using a low-level controller.

Each trial is run for $T_{\max} = 25$ minutes. We consider the number of boxes pushed into the receptacle in this time period as a proxy for our two-tiered objective of maximizing completed boxes and minimizing distance. We tested HeRD and the baseline for 5 trials each. Using the HeRD policy, the robot was able to push 8.0 out of 10 boxes on average into the receptacle in the given time limit, and only 3.2 boxes using the baseline policy.

A sim-to-real gap we noticed was the friction between the boxes and the environment walls. The robot struggled to overcome this force when many boxes were on the wall at once. This was very detrimental to the baseline policy, which relies heavily on gathering many boxes in a corner (sometimes up to seven) and pushing all of them against the wall towards the receptacle. In contrast, the HeRD policy acts more
efficiently,
opting to push one or two boxes in at a time if they are in accessible positions rather than pushing them backwards into a far corner. 

\section{Conclusion}

We introduce HeRD, a hierarchical architecture pairing high-level spatial goal selection with low-level diffusion trajectory generation. This decoupling enables efficient pushing in clutter, consistently outperforming the baseline~\cite{wu_spatial_2020} in success and efficiency, particularly in constrained settings. By leveraging human priors for smooth navigation within a hierarchical framework, HeRD achieves robust generalization across diverse environments.

While currently limited to static 2D environments, HeRD's modular design naturally extends to dynamic settings.  
Future work will explore richer state representations, such as visual observations or expanded object contexts beyond the four nearest boxes, to better capture human priors and improve generalization in complex scenes. 


\bibliographystyle{IEEEtran}
\bibliography{references_clean}

@article{aljalbout_role_2024,
	title = {On the {Role} of the {Action} {Space} in {Robot} {Manipulation} {Learning} and {Sim}-to-{Real} {Transfer}},
	volume = {9},
	issn = {2377-3766},
	doi = {10.1109/LRA.2024.3398428},
	number = {6},
	urldate = {2025-11-03},
	journal = {IEEE Robotics and Automation Letters},
	author = {Aljalbout, Elie and Frank, Felix and Karl, Maximilian and van der Smagt, Patrick},
	month = jun,
	year = {2024},
	pages = {5895--5902},
}

@article{hogan_reactive_2020,
	title = {Reactive planar non-prehensile manipulation with hybrid model predictive control},
	volume = {39},
	issn = {0278-3649},
	doi = {10.1177/0278364920913938},
	language = {EN},
	number = {7},
	urldate = {2025-10-30},
	journal = {The International Journal of Robotics Research},
	author = {Hogan, Francois R and Rodriguez, Alberto},
	month = jun,
	year = {2020},
	pages = {755--773},
}

@inproceedings{vezhnevets_feudal_2017,
	title = {{FeUdal} {Networks} for {Hierarchical} {Reinforcement} {Learning}},
	language = {en},
	urldate = {2025-09-15},
	booktitle = {Proceedings of the 34th {International} {Conference} on {Machine} {Learning}},
	publisher = {PMLR},
	author = {Vezhnevets, Alexander Sasha and Osindero, Simon and Schaul, Tom and Heess, Nicolas and Jaderberg, Max and Silver, David and Kavukcuoglu, Koray},
	month = jul,
	year = {2017},
	pages = {3540--3549},
}

@article{sutton_between_1999,
	title = {Between {MDPs} and semi-{MDPs}: {A} framework for temporal abstraction in reinforcement learning},
	volume = {112},
	issn = {0004-3702},
	shorttitle = {Between {MDPs} and semi-{MDPs}},
	doi = {10.1016/S0004-3702(99)00052-1},
	number = {1},
	urldate = {2025-09-15},
	journal = {Artificial Intelligence},
	author = {Sutton, Richard S. and Precup, Doina and Singh, Satinder},
	month = aug,
	year = {1999},
	pages = {181--211},
}

@article{mnih_playing_2013,
  title={Playing atari with deep reinforcement learning},
  author={Mnih, Volodymyr and Kavukcuoglu, Koray and Silver, David and Graves, Alex and Antonoglou, Ioannis and Wierstra, Daan and Riedmiller, Martin},
  journal={arXiv preprint arXiv:1312.5602},
  year={2013}
}

@article{lin_self-improving_1992,
	title = {Self-improving reactive agents based on reinforcement learning, planning and teaching},
	volume = {8},
	copyright = {1992 Kluwer Academic Publishers},
	issn = {1573-0565},
	doi = {10.1007/BF00992699},
	language = {En},
	number = {3},
	urldate = {2025-09-12},
	journal = {Machine Learning},
	author = {Lin, Long-Ji},
	month = may,
	year = {1992},
	pages = {293--321},
}

@misc{zhong_bench-npin_2025,
	title = {Bench-{NPIN}: {Benchmarking} {Non}-prehensile {Interactive} {Navigation}},
	shorttitle = {Bench-{NPIN}},
	doi = {10.48550/arXiv.2505.12084},
	urldate = {2025-05-20},
	publisher = {arXiv},
	author = {Zhong, Ninghan and Caro, Steven and Iskandar, Avraiem and Ramesh, Megnath and Smith, Stephen L.},
	month = may,
	year = {2025},
    note = {arXiv:2505.12084 [cs]},
}

@inproceedings{wu_spatial_2021,
	title = {Spatial {Intention} {Maps} for {Multi}-{Agent} {Mobile} {Manipulation}},
	doi = {10.1109/ICRA48506.2021.9561359},
	urldate = {2025-09-11},
	booktitle = {{IEEE} {International} {Conference} on {Robotics} and {Automation} ({ICRA})},
	author = {Wu, Jimmy and Sun, Xingyuan and Zeng, Andy and Song, Shuran and Rusinkiewicz, Szymon and Funkhouser, Thomas},
	month = may,
	year = {2021},
	pages = {8749--8756},
}

@article{wu_learning_2022,
	title = {Learning {Pneumatic} {Non}-{Prehensile} {Manipulation} with a {Mobile} {Blower}},
	volume = {7},
	issn = {2377-3766, 2377-3774},
	doi = {10.1109/LRA.2022.3187833},
	number = {3},
	urldate = {2025-09-11},
	journal = {IEEE Robotics and Automation Letters},
	author = {Wu, Jimmy and Sun, Xingyuan and Zeng, Andy and Song, Shuran and Rusinkiewicz, Szymon and Funkhouser, Thomas},
	month = jul,
	year = {2022},
	pages = {8471--8478},
}

@article{ozdamar_pushing_2024,
	title = {Pushing in the {Dark}: {A} {Reactive} {Pushing} {Strategy} for {Mobile} {Robots} {Using} {Tactile} {Feedback}},
	volume = {9},
	issn = {2377-3766},
	shorttitle = {Pushing in the {Dark}},
	doi = {10.1109/LRA.2024.3414279},
	number = {8},
	urldate = {2025-09-11},
	journal = {IEEE Robotics and Automation Letters},
	author = {Ozdamar, Idil and Sirintuna, Doganay and Arbaud, Robin and Ajoudani, Arash},
	month = aug,
	year = {2024},
	pages = {6824--6831},
}

@inproceedings{del_aguila_ferrandis_nonprehensile_2023,
	title = {Nonprehensile {Planar} {Manipulation} through {Reinforcement} {Learning} with {Multimodal} {Categorical} {Exploration}},
	doi = {10.1109/IROS55552.2023.10341629},
	urldate = {2025-09-11},
	booktitle = {{IEEE}/{RSJ} {International} {Conference} on {Intelligent} {Robots} and {Systems} ({IROS})},
	author = {Del Aguila Ferrandis, Juan and Moura, João and Vijayakumar, Sethu},
	month = oct,
	year = {2023},
	pages = {5606--5613},
}

@article{sun_integrating_2023,
	title = {Integrating {Reinforcement} {Learning} and {Learning} {From} {Demonstrations} to {Learn} {Nonprehensile} {Manipulation}},
	volume = {20},
	issn = {1558-3783},
	doi = {10.1109/TASE.2022.3185071},
	number = {3},
	urldate = {2025-09-11},
	journal = {IEEE Transactions on Automation Science and Engineering},
	author = {Sun, Xilong and Li, Jiqing and Kovalenko, Anna Vladimirovna and Feng, Wei and Ou, Yongsheng},
	month = jul,
	year = {2023},
	pages = {1735--1744},
}

@inproceedings{yuan_rearrangement_2018,
	title = {Rearrangement with {Nonprehensile} {Manipulation} {Using} {Deep} {Reinforcement} {Learning}},
	doi = {10.1109/ICRA.2018.8462863},
	urldate = {2025-09-11},
	booktitle = {{IEEE} {International} {Conference} on {Robotics} and {Automation} ({ICRA})},
	author = {Yuan, Weihao and Stork, Johannes A. and Kragic, Danica and Wang, Michael Y. and Hang, Kaiyu},
	month = may,
	year = {2018},
	pages = {270--277},
}

@inproceedings{bauza_data-efficient_2018,
	title = {A {Data}-{Efficient} {Approach} to {Precise} and {Controlled} {Pushing}},
	language = {en},
	urldate = {2025-09-11},
	booktitle = {Proceedings of {The} 2nd {Conference} on {Robot} {Learning}},
	publisher = {PMLR},
	author = {Bauza, Maria and Hogan, Francois R. and Rodriguez, Alberto},
	month = oct,
	year = {2018},
	pages = {336--345},
}

@inproceedings{wu_spatial_2020,
	title = {Spatial {Action} {Maps} for {Mobile} {Manipulation}},
	doi = {10.15607/RSS.2020.XVI.035},
	urldate = {2025-01-16},
	booktitle = {Robotics: {Science} and {Systems} {XVI}},
	author = {Wu, Jimmy and Sun, Xingyuan and Zeng, Andy and Song, Shuran and Lee, Johnny and Rusinkiewicz, Szymon and Funkhouser, Thomas},
	month = jul,
	year = {2020},
}

@article{tang_unwieldy_2023,
	title = {Unwieldy {Object} {Delivery} {With} {Nonholonomic} {Mobile} {Base}: {A} {Stable} {Pushing} {Approach}},
	volume = {8},
	issn = {2377-3766},
	shorttitle = {Unwieldy {Object} {Delivery} {With} {Nonholonomic} {Mobile} {Base}},
	doi = {10.1109/LRA.2023.3322323},
	number = {11},
	urldate = {2025-09-11},
	journal = {IEEE Robotics and Automation Letters},
	author = {Tang, Yujie and Zhu, Hai and Potters, Susan and Wisse, Martijn and Pan, Wei},
	month = nov,
	year = {2023},
	pages = {7727--7734},
}

@article{perez_film_2018,
	title = {{FiLM}: {Visual} {Reasoning} with a {General} {Conditioning} {Layer}},
	volume = {32},
	copyright = {Copyright (c)},
	issn = {2374-3468},
	shorttitle = {{FiLM}},
	doi = {10.1609/aaai.v32i1.11671},
	language = {en},
	number = {1},
	urldate = {2025-09-09},
	journal = {Proceedings of the AAAI Conference on Artificial Intelligence},
	author = {Perez, Ethan and Strub, Florian and Vries, Harm de and Dumoulin, Vincent and Courville, Aaron},
	month = apr,
	year = {2018},
}

@article{wang_hierarchical_2025,
  title={Hierarchical Reinforcement Learning with Uncertainty-Guided Diffusional Subgoals},
  author={Wang, Vivienne Huiling and Wang, Tinghuai and Pajarinen, Joni},
  journal={arXiv preprint arXiv:2505.21750},
  year={2025}
}

@article{wu_diffusion-reinforcement_2025,
  title={Diffusion-reinforcement learning hierarchical motion planning in adversarial multi-agent games},
  author={Wu, Zixuan and Ye, Sean and Natarajan, Manisha and Gombolay, Matthew C},
  journal={arXiv e-prints},
  pages={arXiv--2403},
  year={2024}
}

@inproceedings{ma_hierarchical_2024,
	address = {Seattle, WA, USA},
	title = {Hierarchical {Diffusion} {Policy} for {Kinematics}-{Aware} {Multi}-{Task} {Robotic} {Manipulation}},
	copyright = {https://doi.org/10.15223/policy-029},
	isbn = {979-8-3503-5300-6},
	doi = {10.1109/CVPR52733.2024.01712},
	language = {en},
	urldate = {2025-01-15},
	booktitle = {{IEEE}/{CVF} {Conference} on {Computer} {Vision} and {Pattern} {Recognition} ({CVPR})},
	publisher = {IEEE},
	author = {Ma, Xiao and Patidar, Sumit and Haughton, Iain and James, Stephen},
	month = jun,
	year = {2024},
	pages = {18081--18090},
}

@inproceedings{li_hierarchical_2023,
	title = {Hierarchical {Diffusion} for {Offline} {Decision} {Making}},
	language = {en},
	urldate = {2025-09-08},
	booktitle = {Proceedings of the 40th {International} {Conference} on {Machine} {Learning}},
	publisher = {PMLR},
	author = {Li, Wenhao and Wang, Xiangfeng and Jin, Bo and Zha, Hongyuan},
	month = jul,
	year = {2023},
	pages = {20035--20064},
}

@inproceedings{chen_simple_2024,
  title={Simple Hierarchical Planning with Diffusion},
  author={Chen, Chang and Deng, Fei and Kawaguchi, Kenji and Gulcehre, Caglar and Ahn, Sungjin},
  booktitle={The Twelfth International Conference on Learning Representations},
  year={2024},
  organization={The International Conference on Learning Representations (ICLR)}
}

@inproceedings{janner_planning_2022,
  title={Planning with Diffusion for Flexible Behavior Synthesis},
  author={Janner, Michael and Du, Yilun and Tenenbaum, Joshua and Levine, Sergey},
  booktitle={International Conference on Machine Learning},
  pages={9902--9915},
  year={2022},
  organization={PMLR}
}

@inproceedings{sohl-dickstein_deep_2015,
	title = {Deep {Unsupervised} {Learning} using {Nonequilibrium} {Thermodynamics}},
	language = {en},
	urldate = {2025-09-07},
	booktitle = {Proceedings of the 32nd {International} {Conference} on {Machine} {Learning}},
	publisher = {PMLR},
	author = {Sohl-Dickstein, Jascha and Weiss, Eric and Maheswaranathan, Niru and Ganguli, Surya},
	month = jun,
	year = {2015},
	pages = {2256--2265},
}

@inproceedings{ho_denoising_2020,
	title = {Denoising {Diffusion} {Probabilistic} {Models}},
	volume = {33},
	urldate = {2025-09-02},
	booktitle = {Advances in {Neural} {Information} {Processing} {Systems}},
	publisher = {Curran Associates, Inc.},
	author = {Ho, Jonathan and Jain, Ajay and Abbeel, Pieter},
	year = {2020},
	pages = {6840--6851},
}

@article{fanding_spfa_1994,
	title = {{SPFA} fast algorithm for the shortest path},
	volume = {29},
	number = {2},
	journal = {Journal of Southwest Jiaotong University},
	author = {Fanding, Duan},
	year = {1994},
	pages = {6},
}

@article{stuber_lets_2020,
	title = {Let's {Push} {Things} {Forward}: {A} {Survey} on {Robot} {Pushing}},
	volume = {7},
	issn = {2296-9144},
	shorttitle = {Let's {Push} {Things} {Forward}},
	doi = {10.3389/frobt.2020.00008},
	urldate = {2025-08-08},
	journal = {Frontiers in Robotics and AI},
	author = {Stüber, Jochen and Zito, Claudio and Stolkin, Rustam},
	month = feb,
	year = {2020},
	pages = {8},
}

@inproceedings{hasselt_deep_2015,
  title={Deep reinforcement learning with double q-learning},
  author={Van Hasselt, Hado and Guez, Arthur and Silver, David},
  booktitle={Proceedings of the AAAI conference on artificial intelligence},
  volume={30},
  number={1},
  year={2016}
}

@article{chi_diffusion_2024,
	author = {Cheng Chi and Zhenjia Xu and Siyuan Feng and Eric Cousineau and Yilun Du and Benjamin Burchfiel and Russ Tedrake and Shuran Song},
    title ={Diffusion policy: Visuomotor policy learning via action diffusion},
    journal = {The International Journal of Robotics Research},
    volume = {44},
    number = {10-11},
    pages = {1684-1704},
    year = {2025},
    doi = {10.1177/02783649241273668},
}

\end{document}